\documentclass[a4paper,fleqn]{cas-dc}

\usepackage{bbm}
\usepackage[font=footnotesize,labelfont=rm,textfont=rm]{subfig}
\usepackage[authoryear]{natbib}

\def\tsc#1{\csdef{#1}{\textsc{\lowercase{#1}}\xspace}}
\tsc{WGM}
\tsc{QE}
\tsc{EP}
\tsc{PMS}
\tsc{BEC}
\tsc{DE}

\newcommand{\heatmap}{
    \begin{figure}[t]
      \centering
      \includegraphics[width=0.95\linewidth]{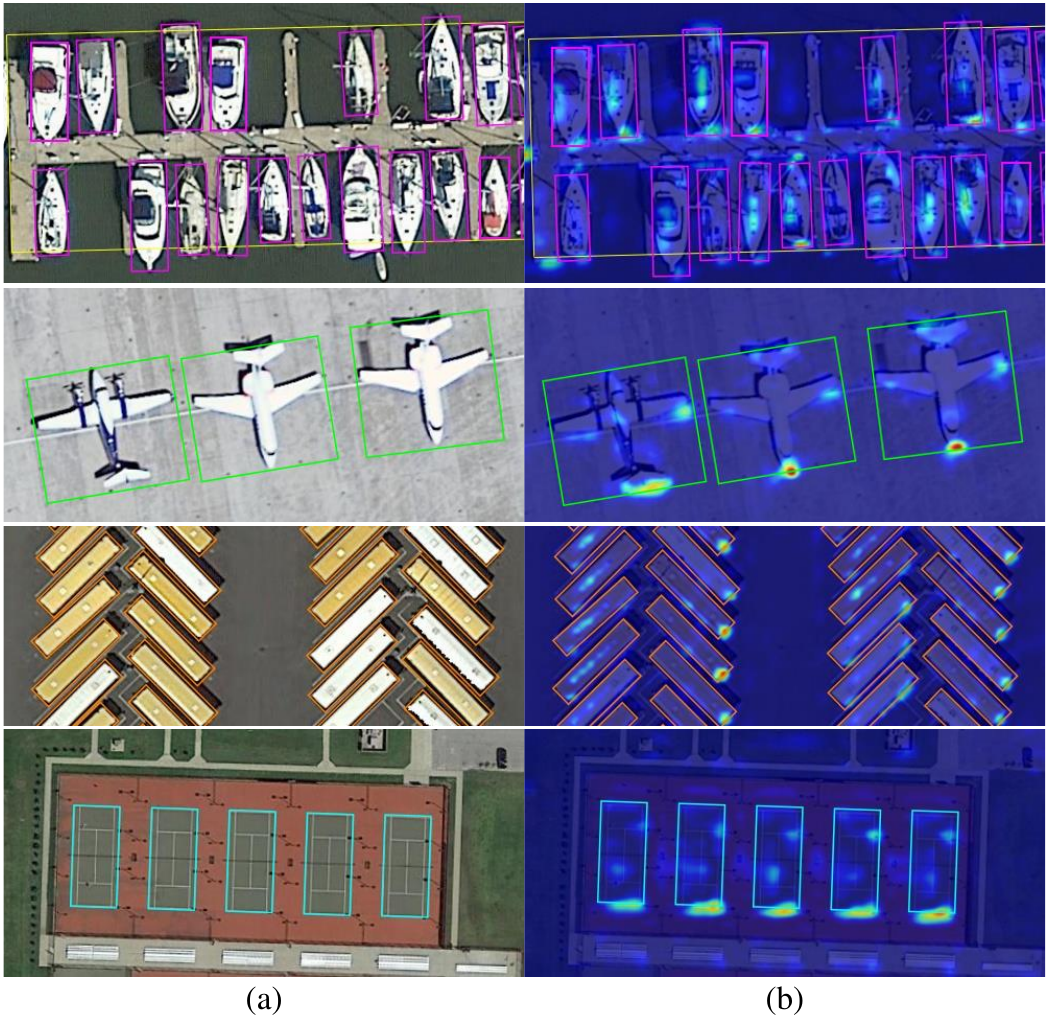}
       \caption{(a) Object instances in oriented object detection exhibit arbitrary orientation, different scales and dense distribution. (b) Visualization of \textbf{RRoI attention} without using CAM. RRoI attention align features and mostly attends to object extremities such as the front and back of ships, planes and vehicles.}
       \label{heat_map}
    \end{figure}
}

\newcommand{\RRoIAttn}{
    \begin{figure*}[!t]
      \centering
       \includegraphics[width=\textwidth]{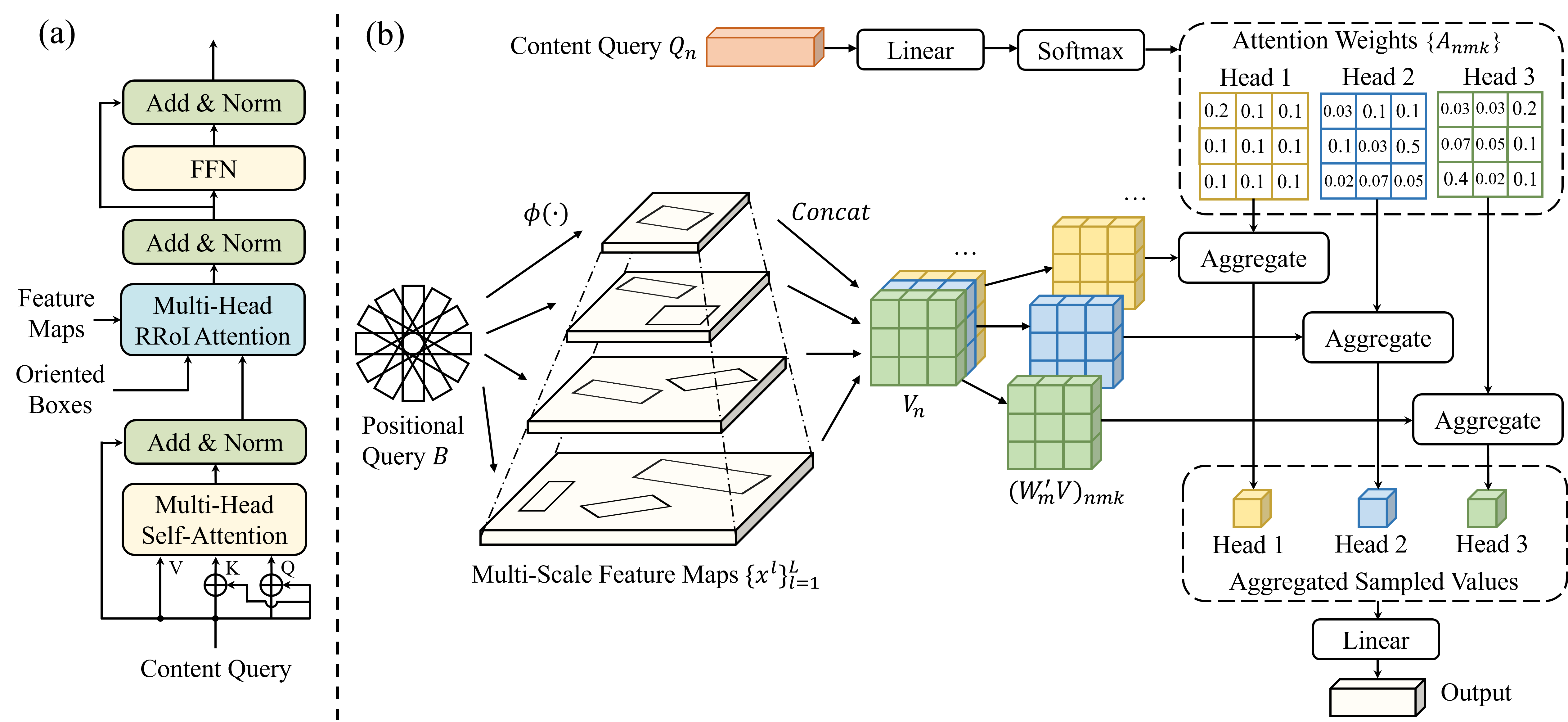}
       \caption{(a) Decoder layer with Rotated RoI attention. (b) Rotated RoI attention.}
       \label{fig_2}
    \end{figure*}
}

\newcommand{\SDQ}{
    \begin{figure}[t]
      \centering
       \includegraphics[width=0.85\linewidth]{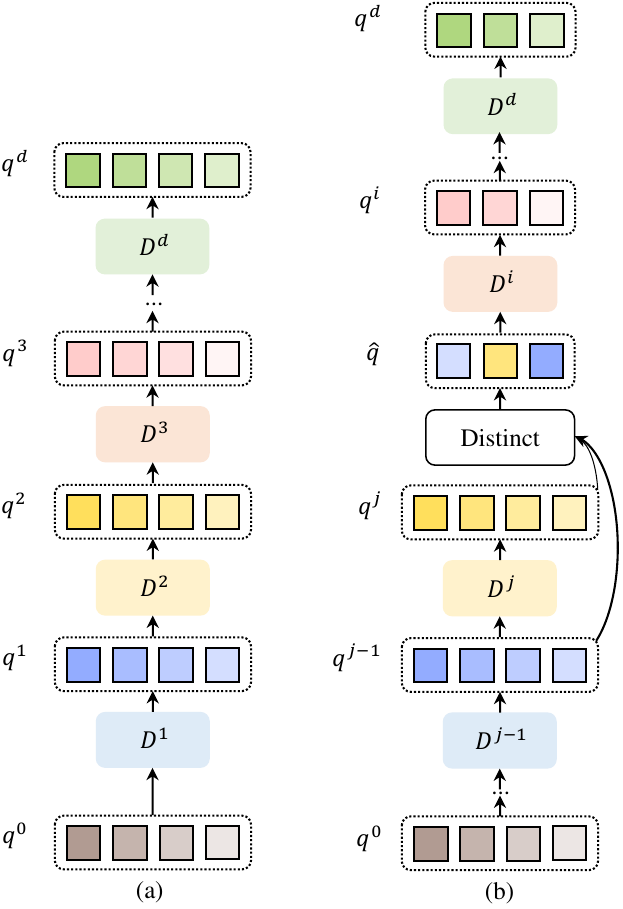}
       \caption{(a) Basic process for queries update layer by layer in decoder. (b) Selective Distinct Queries.}
       \label{fig_3}
    \end{figure}
}

\newcommand{\whysdqwork}{
    \begin{figure}[t]
      \centering
       \includegraphics[width=1.0\linewidth]{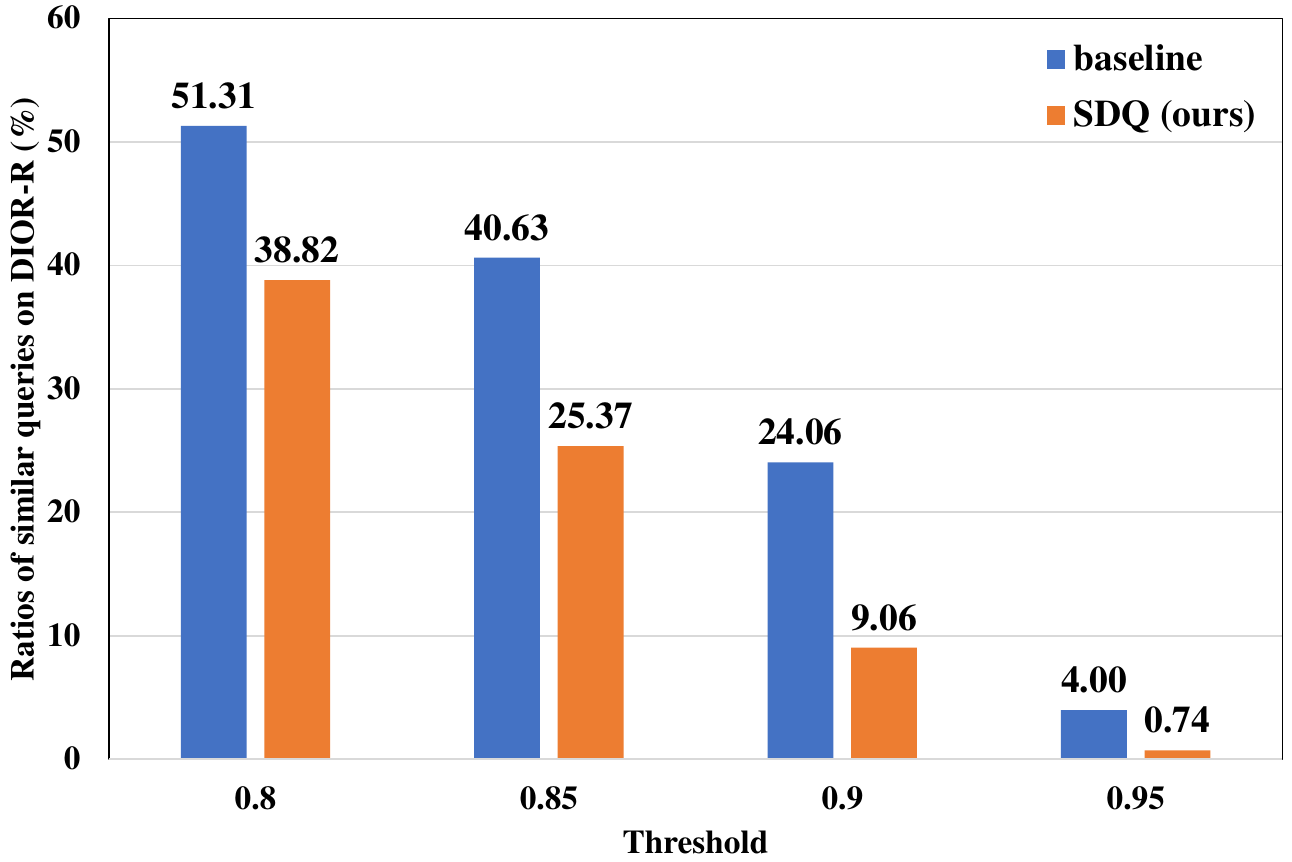}
       \vspace{-4.5mm}
       \caption{The ratios of similar queries under different high IoU threshold on DIOR-R. Compared with baseline, SDQ significantly filters similar queries.}
       \label{why_sdq_work}
    \end{figure}
}

\newcommand{\queries}{
    \begin{figure*}[!t]
      \centering
       \includegraphics[width=1.0\linewidth]{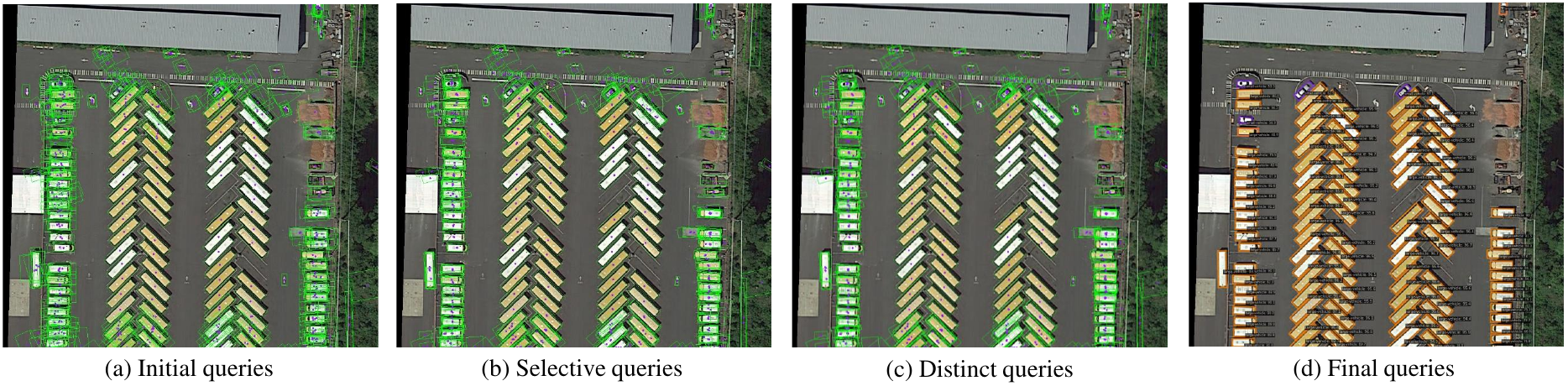}
       \caption{Visualization of different queries in decoder. (a) Initial queries. (b) Selective queries.(c) Distinct queries. (d) Final queries.}
       \label{queries}
    \end{figure*}
}

\newcommand{\DOTAvisual}{
    \begin{figure*}[!t]
      \centering
       \includegraphics[width=1.0\linewidth]{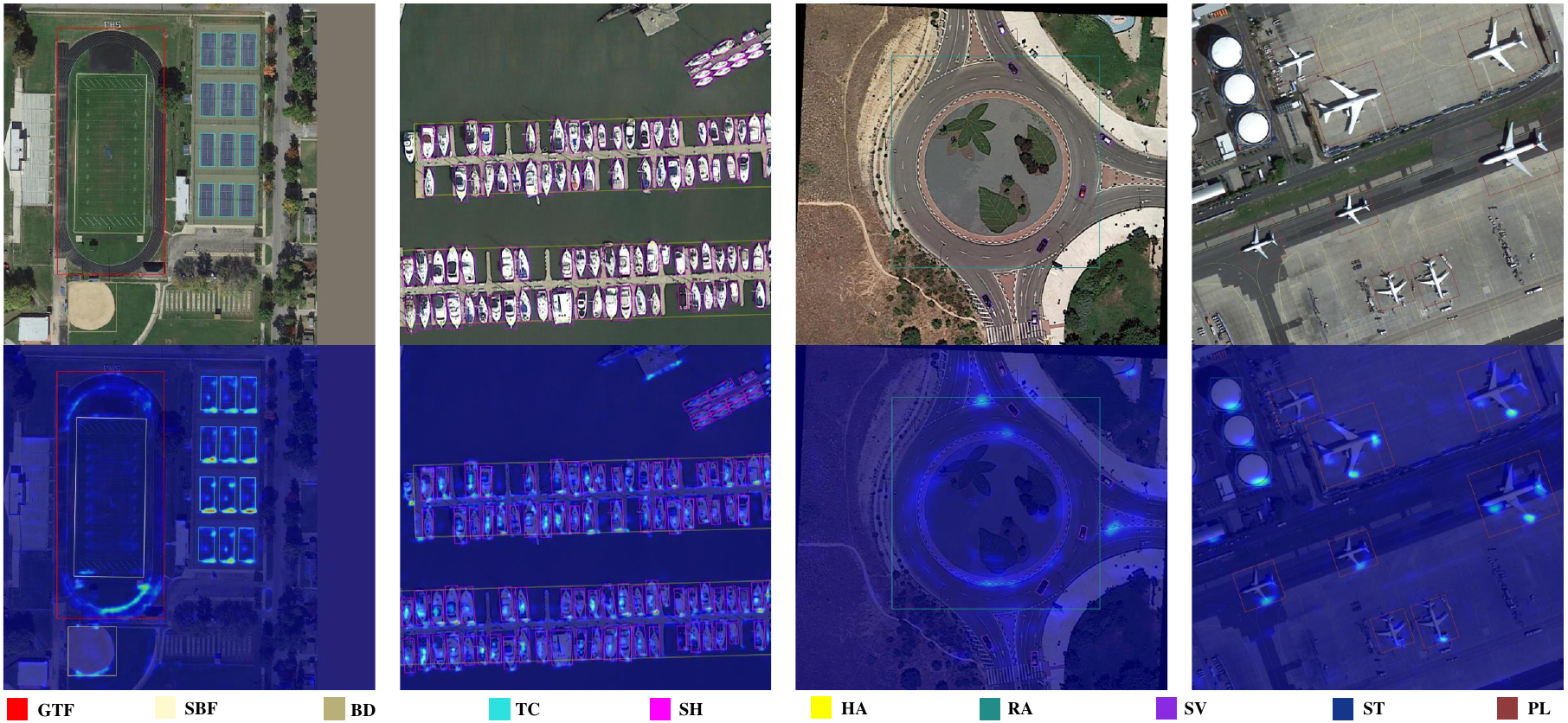}
       \caption{Some detection results on DOTA-v1.0. The left column of images shows the detection results. The right column of images shows RRoI attention.}
       \label{dota_visual}
    \end{figure*}
}

\newcommand{\DIORvisual}{
    \begin{figure*}[!t]
      \centering
       \includegraphics[width=0.99\linewidth]{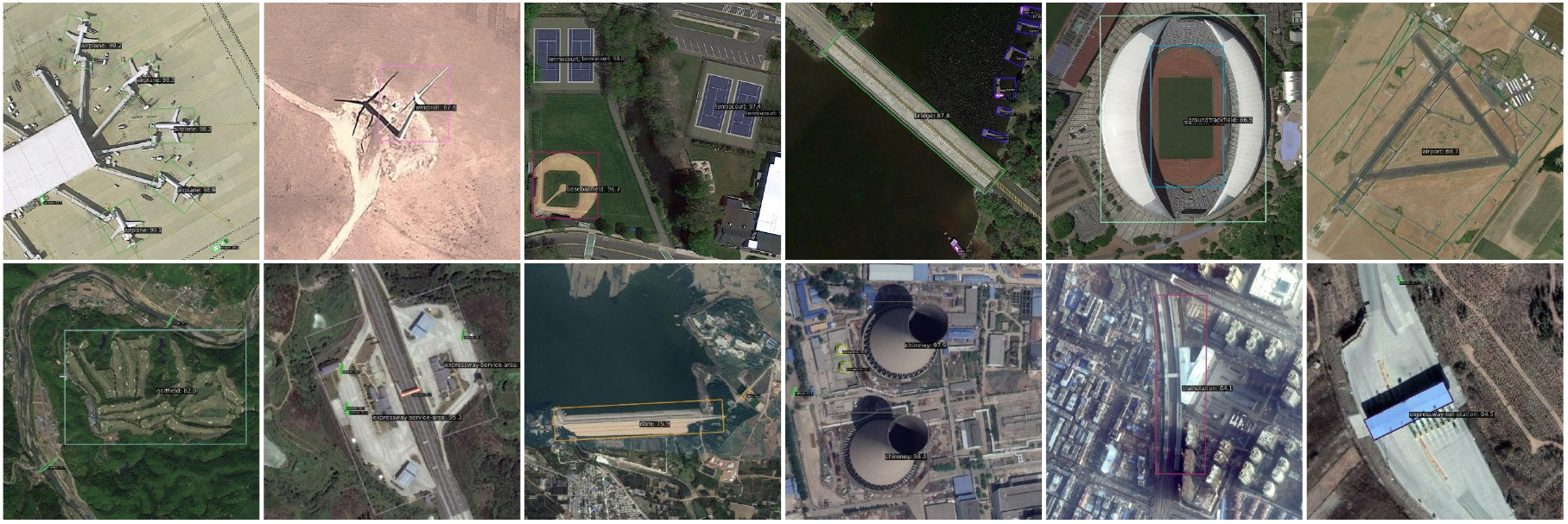}
       \caption{Some detection results on DIOR-R.}
       \label{dior_visual}
    \end{figure*}
}

\newcommand{\overallframework}{
    \begin{figure*}[!t]
      \centering
       \includegraphics[width=\textwidth]{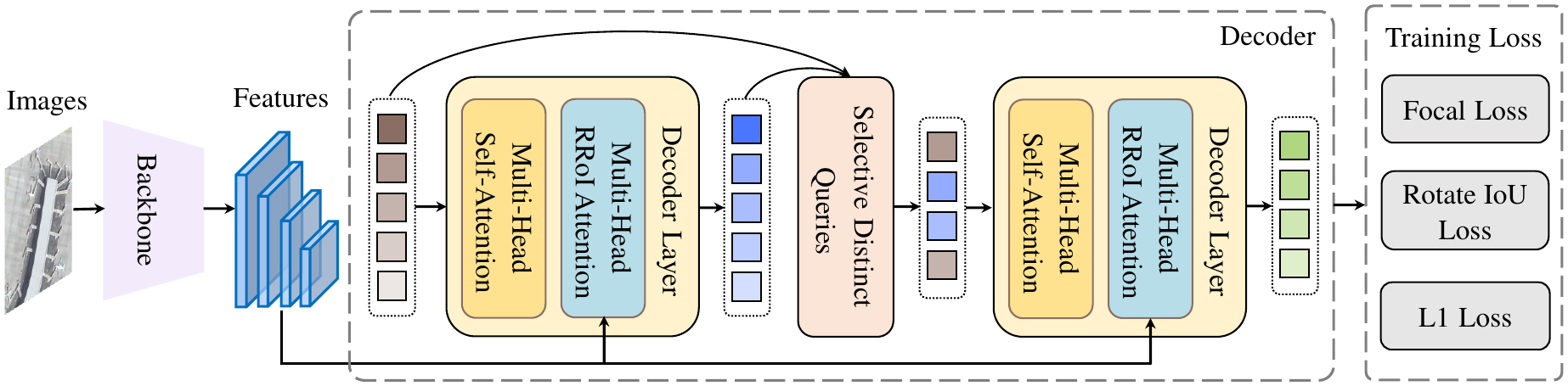}
       \caption{Overall framework. Decoder incorporates Rotated RoI attention (RRoI attention) and Selective Distinct Queries (SDQ). RRoI attention serves as cross-attention, aligning keys with positional queries. SDQ collects queries from intermediate decoder layers and filters similar queries.}
       \label{overall_framework}
    \end{figure*}
}

\newcommand{\convergence}{
    \begin{figure}[!t]
      \centering
       \includegraphics[width=1.0\linewidth]{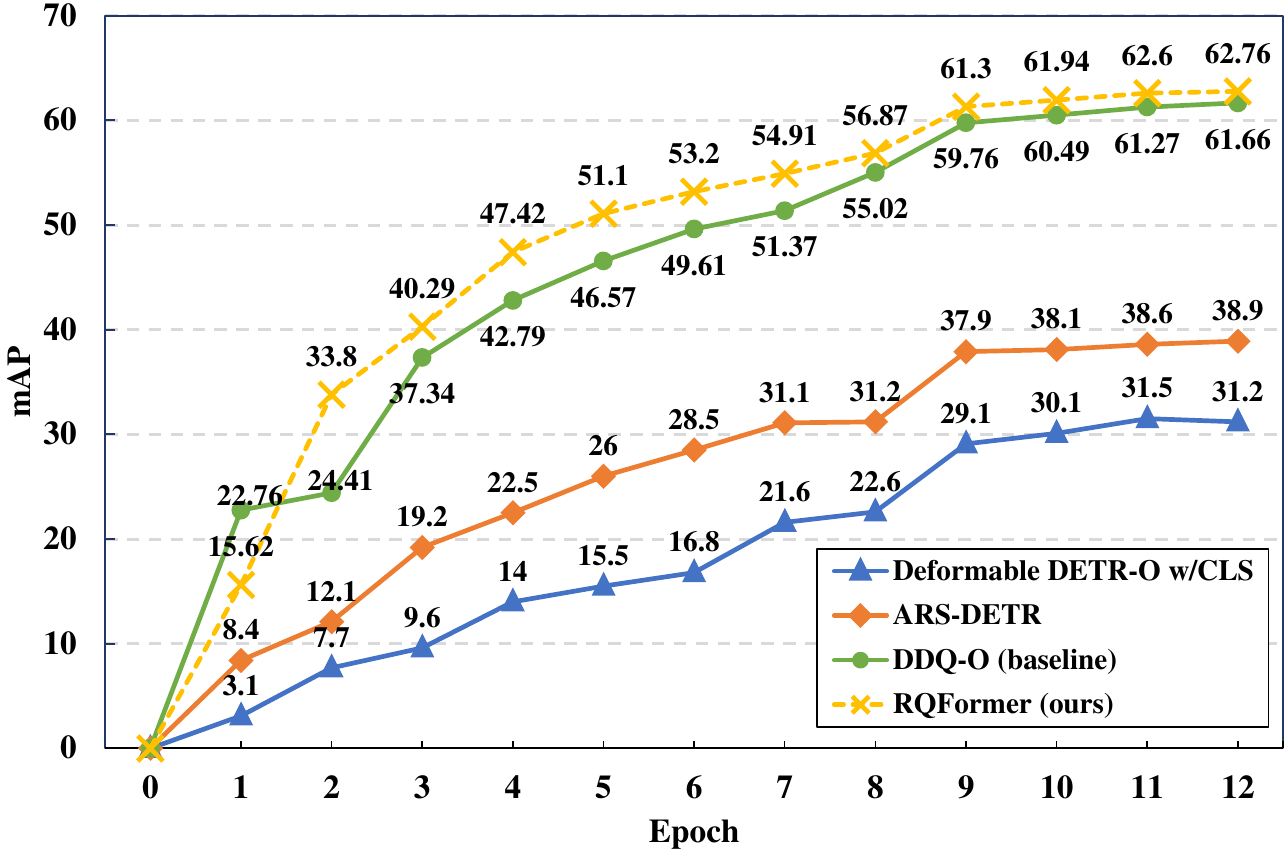}
       \caption{\textbf{Convergence curves} of Deformable DETR-O with CLS, ARS-DETR, DDQ-O and RQFormer with ResNet50 on DIOR-R.}
       \label{convergence_curve}
    \end{figure}
}

\newcommand{\comparisonsimilarquery}{
    \begin{figure}[!t]
      \centering
       \includegraphics[width=1.0\linewidth]{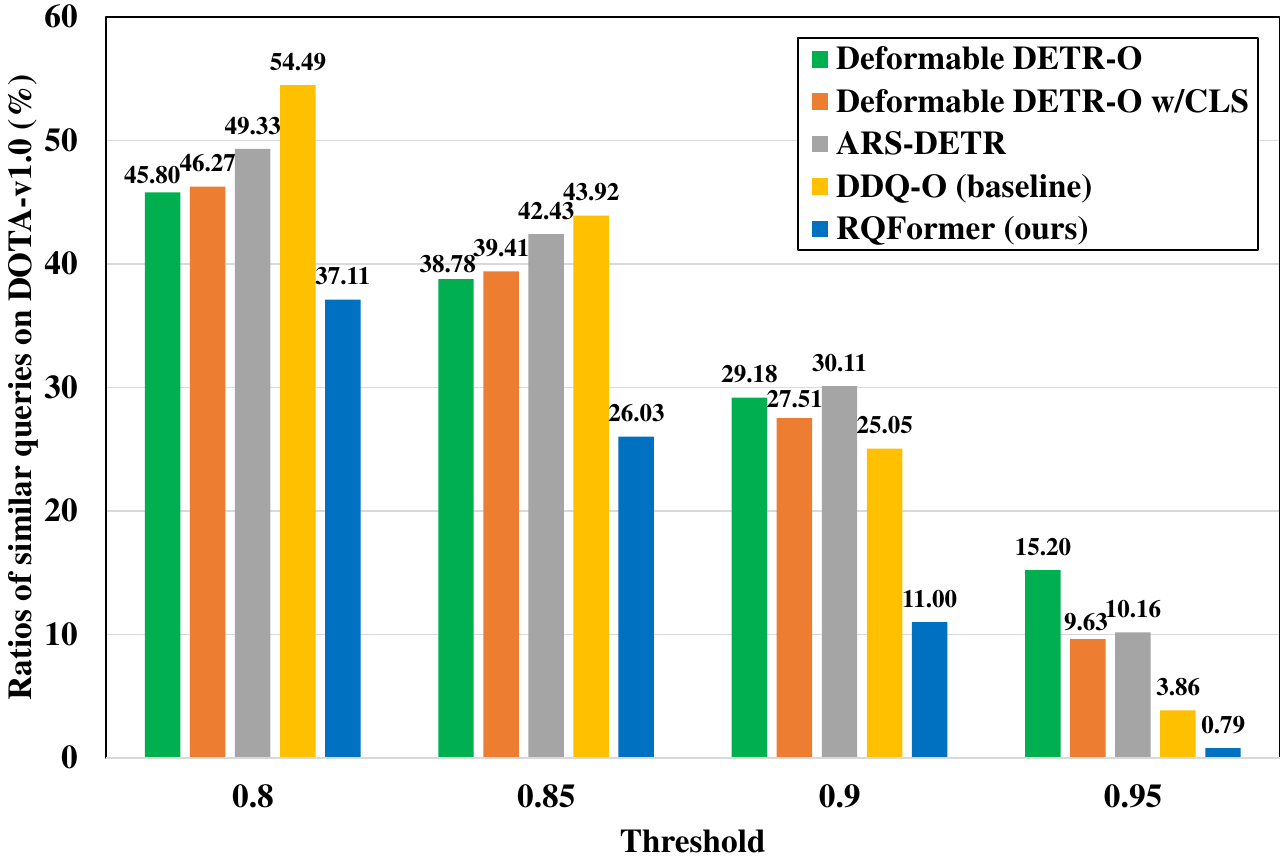}
       \caption{The \textbf{ratios of similar queries} under different high IoU threshold on DOTA-v1.0. Compared with others, our method significantly filters similar queries.}
       \label{comparison_of_similar_query}
    \end{figure}
}

\newcommand{\ICDARvisual}{
    \begin{figure}[!t]
      \centering
       \includegraphics[width=1.0\linewidth]{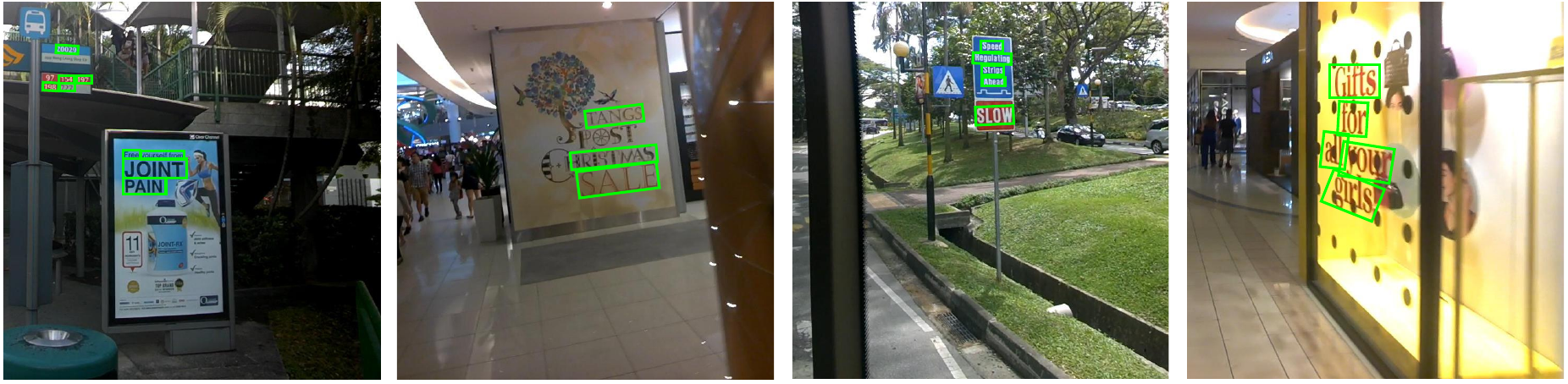}
       \caption{Example results on scene text datasets ICDAR2015.}
       \label{icdar2015_visual}
    \end{figure}
}

\newcommand{\comparewithothermethods}{
    \begin{figure*}[!t]
    \centering
    \subfloat[RQFormers(ours)]
        {
            \includegraphics[width=0.175\linewidth]{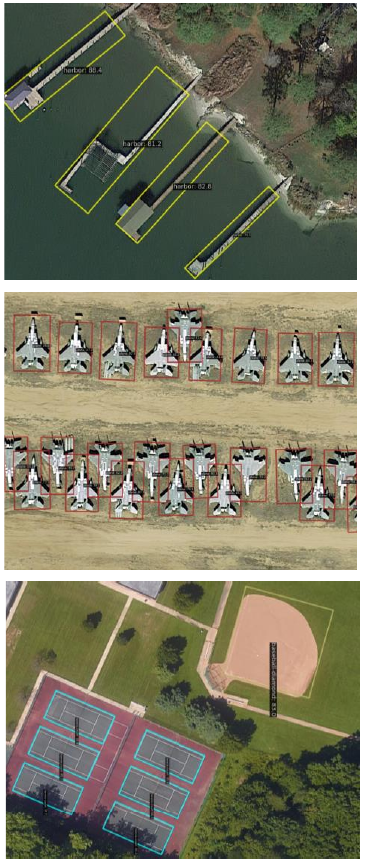}
            \label{compare_orientedformer}
        }
        \hspace{0.5em}
    \subfloat[ARS-DETR]
        {
            \includegraphics[width=0.175\linewidth]{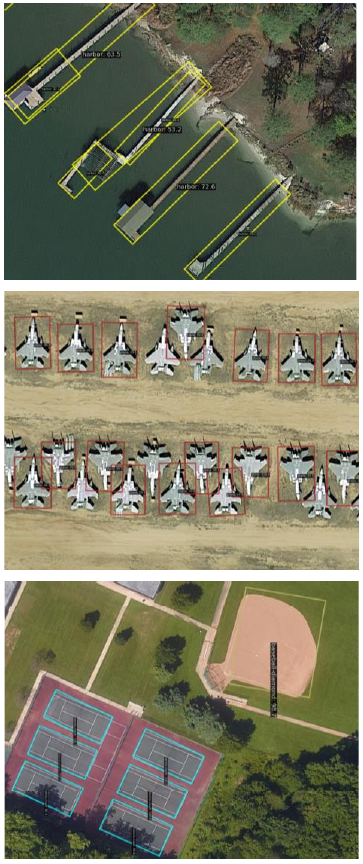}
            \label{compare_orientedrcnn}
        }
        \hspace{0.5em}
    \subfloat[OrientedFormer]
        {
            \includegraphics[width=0.175\linewidth]{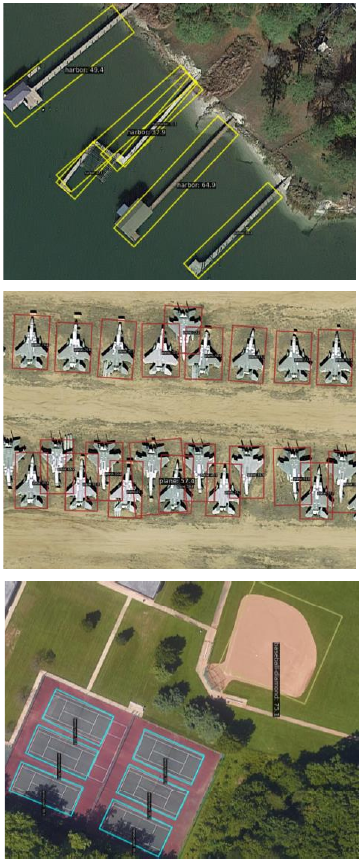}
            \label{compare_sasm}
        }
        \hspace{0.5em}
    \subfloat[Oriented R-CNN]
        {
            \includegraphics[width=0.174\linewidth]{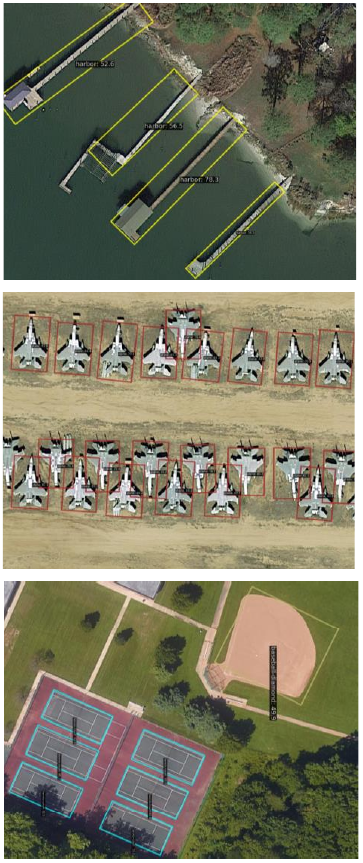}
            \label{compare_arsdetr}
        }
        \hspace{0.5em}
    \subfloat[SASM]
        {
            \includegraphics[width=0.175\linewidth]{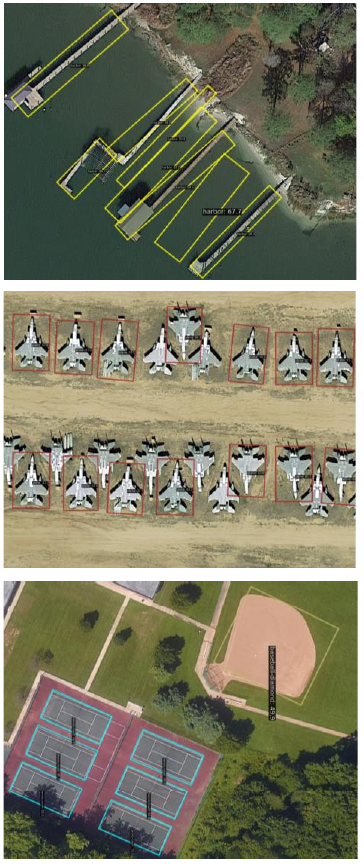}
            \label{compare_deformabledetr}
        }
    \caption{Comparison between our method and others. The confidence threshold is set to 0.3 for better visualization. The 1st line: objects with large aspect ratio, e.g., harbor. The 2nd line: densely packed objects, e.g., planes. The 3rd line: square-like objects, e.g., baseball diamond.}
    \label{compare_with_other_method}
    \end{figure*}

} 
\newcommand{\tablestyle}[2]{\setlength{\tabcolsep}{#1}\renewcommand{\arraystretch}{#2}\centering\footnotesize}

\usepackage{color, xcolor, colortbl}
\definecolor{Gray}{gray}{0.9}

\newcommand{\Tableone}{
    \begin{table}[t]
    \centering
    \renewcommand\arraystretch{1.2}
    \caption{mAP results in different decoder layers. ARS-DETR and Deformable DETR-O with 300 queries are trained on DOTA-v1.0 training and validation set, and the mAP at every decoder layer is reported on test set.}
    \resizebox{\linewidth}{!}
    {
    \begin{tabular}{c|c|c|c|c|c|c}
    \toprule
    Model       & $D^{1}$ & $D^{2}$ & $D^{3}$ & $D^{4}$ & $D^{5}$ & $D^{6}$ \\ \midrule
    ARS-DETR & 0.40    & 3.47   & 19.56   & 33.07   & 39.63   & 72.66   \\
    Deformable DETR-O & 0.13 & 5.49 & 23.64 & 30.00 & 36.61 & 63.42 \\ \bottomrule
    \end{tabular}
    }
    \label{table_1}
    \end{table}
}

\newcommand{\TableDIORResult}{
    \begin{table*}[t]\huge
    \renewcommand\arraystretch{1.1}
    \caption{Comparison with state-of-the-art methods on the \textbf{DIOR-R}. The results in \textbf{bold} denote the best performance of each column.}
    \resizebox{\linewidth}{!}
    {
    \begin{tabular}{l|cccccccccccccccccccc|c}
    \toprule
    Mehtod              & APL            & APO            & BF             & BC             & BR             & CH             & DAM            & ETS            & ESA            & GF             & GTF            & HA             & OP             & SH             & STA            & STO            & TC             & TS             & VE             & WM             & mAP            \\ \midrule
    \textit{one-stage:}  &                &                &                &                &                &                &                &                &                &                &                &                &                &                &                &                &                &                &                &                &                \\
    RetinaNet-O          & 61.49          & 28.52          & 73.57          & 81.17          & 23.98          & 72.54          & 19.94          & 72.39          & 58.20          & 69.25          & 79.54          & 32.14          & 44.87          & 77.71          & 67.57          & 61.09          & 81.46          & 47.33          & 38.01          & 60.24          & 57.55          \\
    DFDet & 61.92 & 38.83 & 77.41 & 81.36 & 34.11 & 74.94 & 26.26 & 62.31 & 76.06 & 75.56 & 79.62 & 38.26 & 52.76 & 80.40 & 73.11 & 68.27 & 81.38 & 52.23 & 44.11 & 63.35 & 62.11 \\
    SOOD & 62.61 & 46.82 & 72.00 & 81.28 & 41.40 & 72.56 & 78.63 & 68.91 & 27.53 & 78.64 & 81.35 & 46.88 & 56.32 & 80.57 & 73.74 & 68.36 & 81.43 & 56.91 & 47.33 & 65.52 & 64.41 \\
    Oriented Rep & \textbf{70.03} & 46.11 & 76.12  & 87.19 & 39.14 & \textbf{78.76}  & 34.57  & 71.80& 80.42 & 76.16 & 79.41  & 45.48  & 54.90 & 87.82  & 77.03 & 68.07 & 81.60  & 56.83 & 51.57 & \textbf{71.25} & 66.71 \\
    DCFL                 & 68.60          & 53.10          & 76.70 & 87.10          & 42.10          & 78.60 & 34.50          & 71.50          & 80.80          & \textbf{79.70} & 79.50          & 47.30          & 57.40          & 85.20          & 64.60          & 66.40          & 81.50          & 58.90          & 50.90          & 70.90          & 66.80          \\ \midrule
    \textit{two-stage:}  &                &                &                &                &                &                &                &                &                &                &                &                &                &                &                &                &                &                &                &                &                \\
     Faster RCNN-O & 62.79          & 26.80          & 71.72          & 80.91          & 34.20          & 72.57          & 18.95          & 66.45          & 65.75          & 66.63          & 79.24          & 34.95          & 48.79          & 81.14          & 64.34          & 71.21          & 81.44          & 47.31          & 50.46          & 65.21          & 59.54          \\
    Gliding Vertex       & 65.35          & 28.87          & 74.96          & 81.33          & 33.88          & 74.31          & 19.58          & 70.72          & 64.70          & 72.30          & 78.68          & 37.22          & 49.64          & 80.22          & 69.26          & 61.13          & 81.49          & 44.76          & 47.71          & 65.04          & 60.06          \\
    RoI Transformer      & 63.34          & 37.88          & 71.78          & 87.53          & 40.68          & 72.60          & 26.86          & 78.71          & 68.09          & 68.96          & 82.74          & 47.71          & 55.61          & 81.21          & \textbf{78.23} & 70.26          & 81.61          & 54.86          & 43.27          & 65.52          & 63.87          \\
    QPDet                & 63.22          & 41.39          & 71.97          & \textbf{88.55} & 41.23          & 72.63          & 28.82          & \textbf{78.90} & 69.00          & 70.07          & \textbf{83.01} & \textbf{47.83} & 55.54          & 81.23          & 72.15          & 62.66          & \textbf{89.05} & 58.09          & 43.38          & 65.36          & 64.20          \\
    AOPG                 & 62.39          & 37.79          & 71.62          & 87.63          & 40.90          & 72.47          & 31.08          & 65.42          & 77.99          & 73.20          & 81.94          & 42.32          & 54.45          & 81.17          & 72.69          & 71.31          & 81.49          & 60.04 & 52.38          & 69.99          & 64.41          \\ \midrule
    \textit{end-to-end:} &                &                &                &                &                &                &                &                &                &                &                &                &                &                &                &                &                &                &                &                &                \\
    ARS-DETR             & 65.82          & 53.40          & 74.22          & 81.11          & 42.13          & 76.23          & 38.90          & 71.52          & \textbf{82.24} & 75.91          & 77.91          & 33.03          & 57.02          & 84.82          & 69.71          & 72.20          & 80.33          & 58.91          & 51.52          & 70.73          & 65.90          \\ 
    OrientedFormer& 65.65& 48.69 & \textbf{78.79} & 87.17 & 41.90 & 76.34 & 34.37 & 72.14 & 81.40 & 75.34 & 79.83 & 45.15 & 56.12 & 88.66& 67.59 & 72.68 & 87.32 & \textbf{60.31} & 56.54 & 69.56 & 67.28 \\
    \midrule
    \textit{end-to-end:} &                &                &                &                &                &                &                &                &                &                &                &                &                &                &                &                &                &                &                &                &                \\
    DDQ-O (baseline) & 66.71 & 54.14 & 73.11 & 81.80 & \textbf{45.25} & 77.60 & 33.37 & 70.09 & 79.18 & 72.24 & 75.42 & 47.18 & 58.27 & \textbf{89.50} & 70.98 & 74.62 & 82.83 &55.03& \textbf{56.23} & 66.61 & 66.51 \\
    RQFormer           & 67.31          & \textbf{55.23} & 74.19          & 82.74          & 44.49 & 78.56          & \textbf{39.85} & 70.27          & 79.84          & 75.10          & 80.38          & 45.64          & \textbf{58.51} & 88.91 & 68.10          & \textbf{75.73} & 85.52          & 57.17          & 53.54 & 65.05          & \textbf{67.31} \\ \bottomrule
    \end{tabular}
    }

    \label{DIOR-R-result}
    \end{table*}
}

\newcommand{\TableDOTAVoneResult}{
    \begin{table*}[t] \large
    \renewcommand\arraystretch{1.1}
    \caption{Comparison with state-of-the-art methods on the \textbf{DOTA-v1.0} dataset. * denotes multi-scale training and testing. \emph{D. DETR} represents Deformable DETR. The results in \textbf{bold} denote the best performance of each column.}
    \resizebox{\linewidth}{!}
    {
    \begin{tabular}{l|c|ccccccccccccccc|c}
    \toprule
    Method   & Backbone & PL    & BD    & BR    & GTF   & SV    & LV    & SH    & TC    & BC    & ST    & SBF   & RA    & HA    & SP    & HC    & mAP   \\ \midrule
    \textit{one-stage:}  & & & & & & & & & & & & & & & & & \\
    PSC      &   R50      & 88.24 & 74.42 & 48.63 & 63.44 & 79.98 & 80.76 & 87.59 & 90.88 & 82.02 & 71.58 & 59.12 & 60.78 & 65.78 & 71.21 & 53.06 & 71.83 \\
    TIR-Net &  R50 & 88.95 & 74.48 & 48.96 & 59.60 & 80.18 & 80.64 & 87.85 & 90.89 & 83.74 & 85.69 & 62.40 & 63.40 & 67.41 & 70.55& 45.86 & 72.71\\
    R3Det      & R101     & 88.76 & 83.09 & 50.91 & 67.27 & 76.23 & 80.39 & 86.72 & 90.78 & 84.68 & 83.24 & 61.98 & 61.35 & 66.91 & 70.63 & 53.94 & 73.79 \\
    CFA                  & R50      & 88.34 & 83.09 & 51.92 & 72.23 & 79.95 & 78.68 & 87.25 & 90.90 & 85.38 & 85.71 & 59.63 & 63.05 & 73.33 & 70.36 & 47.86 & 74.51 \\
    SASM           & R50      & 86.42 & 78.97 & 52.47 & 69.84 & 77.30 & 75.99 & 86.72 & 90.89 & 82.63 & 85.66 & 60.13 & 68.25 & 73.98 & 72.22 & 62.37 & 74.92 \\
    \midrule
    \textit{two-stage:}  &          &       &       &       &       &       &       &       &       &       &       &       &       &       &       &       &       \\
    H2RBox & R50      & 88.24 & 79.30 & 42.76 & 55.79 & 78.90 & 72.70 & 77.54 & 90.85 & 81.96 & 84.38 & 55.28 & 64.49 & 61.91 & 70.63 & 51.51 & 70.41 \\ 
    SCRDet   & R101     & \textbf{89.98} & 80.65 & 52.09 & 68.36 & 68.36 & 60.32 & 72.41 & 90.85 & 87.94 & 86.86 & 65.02 & 66.68 & 66.25 & 68.24 & 65.21 & 72.61 \\
    RoI Transformer      & R50      & 88.65 & 82.60 & 52.53 & 70.87 & 77.93 & 76.67 & 86.87 & 90.71 & 83.83 & 52.81 & 53.95 & 67.61 & 74.67 & 68.75 & 61.03 & 74.61 \\
    Gliding Vertex       & R101     & 89.64 & 85.00 & 52.26 & 77.34 & 73.01 & 73.14 & 86.82 & 90.74 & 79.02 & 86.81 & 59.55 & \textbf{70.91} & 72.94 & 70.86 & 57.32 & 75.02 \\
    Oriented R-CNN     & R50      & 89.46 & 82.12 & 54.78 & 70.86 & 78.93 & 83.00 & 88.20 & 90.90 & 87.50 & 84.68 & 63.97 & 67.69 & 74.94 & 68.84 & 52.28 & 75.87 \\ \midrule
    \textit{end-to-end:} &          &       &       &       &       &       &       &       &       &       &       &       &       &       &       &       &       \\
    D. DETR-O   & R50      & 78.95 & 68.64 & 32.57 & 55.17 & 72.53 & 57.77 & 73.71 & 88.36 & 75.46 & 79.34 & 45.36 & 53.78 & 52.94 & 66.35 & 50.38 & 63.42 \\
    D. DETR-O w/ CSL     & R50      & 86.27 & 76.66 & 46.64 & 65.29 & 76.80 & 76.32 & 87.74 & 90.77 & 79.38 & 82.36 & 54.00 & 61.47 & 66.05 & 70.46 & 61.97 & 72.15 \\
    EMO2-DETR & R50 & 88.08 & 77.91 & 43.17 & 62.91 & 74.01 & 75.09 & 79.21 & 90.88 & 81.50 & 84.04 & 51.92 & 59.44 & 64.74 & 71.81 & 58.96 & 70.91 \\
    ARS-DETR          & R50      & 86.61 & 77.26 & 48.84 & 66.76 & 78.38 & 78.96 & 87.40 & 90.61 & 82.76 & 82.19 & 54.02 & 62.61 & 72.64 & 72.80 & 64.96 & 73.79 \\  \midrule
    \textit{end-to-end:} &          &       &       &       &       &       &       &       &       &       &       &       &       &       &       &       &       \\
    RQFormer   & R50      & 87.45 & 78.57 & 47.36 & 69.01 & 79.58 & 81.27 & 88.53 & 90.89 & 82.80 & 86.21 & 58.68 & 64.20 & 75.21 & 74.44 & 61.39 & 75.04 \\
    RQFormer*    & R50      & 87.98 & \textbf{86.13} & \textbf{57.79} & \textbf{82.12} & \textbf{81.33} & \textbf{84.87} & \textbf{88.78} & \textbf{90.90} & \textbf{88.65} & \textbf{87.45} & \textbf{70.74} & 70.39 & \textbf{80.33} & \textbf{81.03} & \textbf{75.31} & \textbf{80.92} \\ \bottomrule
    \end{tabular}
    }

    \label{DOTA-1.0-result}
    \end{table*}
}

\newcommand{\TableDOTAVonefiveResult}{
    \begin{table}[!htbp]
    \renewcommand\arraystretch{1.1}
    \caption{Main results of small size objects on \textbf{DOTA-v1.5}. The results in \textbf{bold} denote the best performance of each column.}
    \resizebox{\linewidth}{!}
    {

    \begin{tabular}{l|cccccc|c}
    \toprule
    Method        & SV    & LV    & SH    & ST    & SP    & CC    & mAP   \\ \midrule
    RetinaNet-O   & 44.53 & 56.79 & 73.31 & 59.96 & 64.52 & 0.83  & 59.16 \\
    DCFL          & 56.72 & -     & 80.87 & 75.65 & -     & -     & 67.37 \\
    Faster RCNN-O & 51.28 & 68.98 & 79.37 & 67.50 & 65.28 & 1.54  & 62.00 \\
    Mask R-CNN    & 51.31 & 71.34 & 79.75 & 66.07 & 64.46 & 9.42  & 62.67 \\
    HTC           & 51.54 & 73.31 & 80.31 & 67.34 & 64.48 & 5.15  & 63.40 \\
    ReDet         & 52.38 & 75.73 & 80.92 & 68.64 & 70.55 & 11.53 & 66.86 \\
    RQFormer    & \textbf{62.16} & \textbf{78.41} & \textbf{88.94} & \textbf{81.83} & \textbf{72.69} & \textbf{13.37} & \textbf{67.43} \\ \bottomrule
    \end{tabular}

    }

    \label{DOTA-1.5-result}
    \end{table}
}

\newcommand{\TableDOTAVtwoResult}{
    \begin{table}[!htbp]\large
    \renewcommand\arraystretch{1.1}
    \caption{Performance comparisons on the \textbf{DOTA-v2.0} dataset.}
    \resizebox{\linewidth}{!}
    {

    \begin{tabular}{c|cccc}
    \toprule
    Method & SASM            & RetinaNet-O    & Oriented Rep   & Mask R-CNN \\
    mAP    & 44.53           & 46.68          & 48.95          & 49.47      \\ \midrule
    Method & ATSS-O          & S$^{2}$A-Net        & HTC            & DCFL       \\
    mAP    & 49.57           & 49.86          & 50.34          & 51.57      \\ \midrule
    Method & RoI Trans. & S$^{2}$A-Net$+$DCFL & Oriented R-CNN & RQFormer \\
    mAP    & 52.81           & 52.84          & 53.28          & \textbf{53.28}      \\ \bottomrule
    \end{tabular}
    
    }

    \label{DOTA-2.0-result}
    \end{table}
}

\newcommand{\TableSDQSparsercnn}{
    \begin{table}[!t]
    \centering
    \Large
    \renewcommand\arraystretch{1.22}
    \caption{Evaluation Results on \textbf{MS COCO 2017}. The proposed SDQ is applied to horizontal object detector Sparse R-CNN and DDQ.}
    \resizebox{\linewidth}{!}
    {
    \begin{tabular}{c|ccc|lcc|cc}
    \toprule
    Model                                                                                     & w/SDQ     & query   & epochs   & mAP                          & AP$_{50}$    & AP$_{75}$  & FPS  & Params\\ 
    \midrule
    \multirow{4}{*}{\begin{tabular}[c]{@{}c@{}}Sparse\\ R-CNN\\ \end{tabular}} &  & 100     & 12       & 37.9                         & 56.0    & 40.5    & 20.6 & 106M      \\
    &\checkmark & 100     & 12       & \textbf{41.0} ($\uparrow$ \textbf{3.1}) & 59.4    & 44.4    & 20.0 & 106M    \\
    &           & 300     & 36       & 45.0                         & 64.1    & 48.9    & 20.5 & 106M      \\
    &\checkmark & 300     & 36       & \textbf{46.8} ($\uparrow$ \textbf{1.8}) & 65.9    & 51.2    & 19.3 & 106M    \\
    \midrule
    \multirow{2}{*}{\begin{tabular}[c]{@{}c@{}}DDQ\\ \end{tabular}} &  & 100     & 12  & 42.1                                   & 59.8 & 46.3 & 9.5 & 64M \\
    & \checkmark & 100     & 12  & \textbf{42.4} ($\uparrow$ \textbf{0.3})& 59.6 & 46.5 & 9.4 & 64M \\
    \bottomrule
    \end{tabular}
    }

    \label{SDQ_result}
    \end{table}
}

\newcommand{\TableIndividualStrategy}{

    \begin{tabular}{c|cc|ccc}
    \toprule
    Method                      & RRoI Att. & SDQ & mAP   & Params(M) & FPS  \\ \midrule
    DDQ-O(baseline)        &           &     & 61.66 & 63.68        &14.40  \\ \midrule
    RQFormer &\checkmark &           & 62.31 & 40.88  & 14.76 \\
    RQFormer &           &\checkmark & 62.41 & 63.68  & 14.12 \\
    \rowcolor{Gray}
    RQFormer &\checkmark &\checkmark & 62.76 & 40.88  & 14.28 \\ \bottomrule
    \end{tabular}

}

\newcommand{\TableAttentionHead}{

    \newcolumntype{a}{>{\columncolor{Gray}}c}
    \resizebox{1.0\linewidth}{!}{
        \begin{tabular}{c|ccac}
        \toprule
        Head        & 2     & 4     & 8     & 16    \\ \midrule
        mAP         & 61.36 & 62.19 & 62.63 & 55.63 \\
        Params(M)   & 40.73 & 40.78 & 40.88 & 41.08 \\ \bottomrule
        \end{tabular}
    }

}

\newcommand{\Tablepoolingsize}{

    \newcolumntype{a}{>{\columncolor{Gray}}c}
    \resizebox{1.0\linewidth}{!}{
        \begin{tabular}{c|ccac}
        \toprule
        Pooling Size & 3     & 5     & 7     & 9     \\ \midrule
        mAP          & 61.02 & 61.35 & 62.63 & 62.32 \\
        Params(M)    & 40.71 & 40.78 & 40.88 & 41.01 \\ \bottomrule
        \end{tabular}
    }

}

\newcommand{\TableIOUthreshold}{

    \renewcommand\arraystretch{1.53}
    \newcolumntype{a}{>{\columncolor{Gray}}c}
    \resizebox{1.03\linewidth}{!}{
        \begin{tabular}{c|cacc}
        \toprule
        threshold & 0.8   & 0.85  & 0.9   & 0.95  \\ \midrule
        mAP       & 62.58 & 62.76 & 62.63 & 62.49 \\
        Params(M) & 40.88 & 40.88 & 40.88 & 40.88 \\ \bottomrule
        \end{tabular}
    }

}

\newcommand{\TabledecoderlayerinSDQ}{

        \begin{tabular}{c|c|c|c|c c}
        \toprule
        Method               & Layers              & Selective & Distinct & mAP & Params(M)  \\ \midrule
        SDQ & 3 & $0^{th}$, $1^{st}$, $2^{nd}$   & $2^{nd}$      & 58.01 &42.70 \\
        SDQ & 3 & $1^{st}$, $2^{nd}$      & $2^{nd}$      & 60.66 &42.70\\
        SDQ & 3 & $1^{st}$         & $1^{st}$      & 58.03 &42.70\\
        \rowcolor{Gray}
        SDQ & 3 & $2^{nd}$         & $2^{nd}$      & 62.08 &42.70\\ \bottomrule
        \end{tabular}
}

\newcommand{\TableReduceDecoderLayer}{

    \resizebox{1.0\linewidth}{!}{
        \begin{tabular}{c|c|c|c|c}
        \toprule
        Method               & Layers & Selective & Distinct & mAP    \\ \midrule
        SDQ & 3      & $2^{nd}$         & $2^{nd}$      & 62.08 \\
        \rowcolor{Gray}
        SDQ & 2      & $1^{st}$         & $1^{st}$      & 62.63 \\
        SDQ & 1      & $0^{th}$         & $0^{th}$      & 44.02 \\ \bottomrule
        \end{tabular}
    }

}

\newcommand{\TableQuerySelectionMechanism}{

    \resizebox{1.0\linewidth}{!}{
        \begin{tabular}{c|c|c|c|c}
        \toprule
        Method & Layers              & Selective                 & Distinct               & mAP    \\ \midrule
        Top-$n$  & 3 & $1^{st}$, $2^{nd}$ & $2^{nd}$ & 59.49 \\
        Stack  & 3 & $1^{st}$, $2^{nd}$ & $2^{nd}$ & 59.99 \\
        \rowcolor{Gray}
        SDQ    & 3 & $1^{st}$, $2^{nd}$ & $2^{nd}$ & 60.66 \\ \bottomrule
        \end{tabular}
    }

}

\newcommand{\TableAttentionMechanism}{

        \begin{tabular}{c|ccc}
        \toprule
        Attention     & mAP   & Params(M) & FPS  \\ \midrule
        Instance interact & 62.41 & 63.68     & 14.12\\
        Dynamic Attention  & 61.59 & 46.97     & 11.70 \\
        \rowcolor{Gray}
        RRoI Attention     & 62.76 & 40.88     & 14.28\\ \toprule
        \end{tabular}

}
\newcommand{\TableResultICDAR}{
    \begin{table}[!htbp]
    \centering
    \renewcommand\arraystretch{1.1}
    \caption{Comparison of the performance of different methods on \textbf{ICDAR2015}. 'P' is precision and 'R' is recall. The results in \textbf{bold} denote the best performance of each column. The backbone used by all methods is Resnet50.}
    \resizebox{\linewidth}{!}
    {
        \begin{tabular}{l|cccc}
        \toprule
        Method        & P    & R    & F-measure & FPS  \\ \midrule
        R3DET & 71.7 & 79.8 & 75.5      & -    \\
        GWD & 74.0 & \textbf{80.5} & 77.1      & -    \\
        PSC  & 84.2 & 73.9 & 78.7      & 20.4 \\
        Oriented Reppoints & 84.3 & 74.0 & 78.8      & 21.4 \\
        CFA & \textbf{85.3} & 73.5 & 78.9      & \textbf{21.4} \\
        S$^{2}$A-Net & 80.4 & 78.2 & 79.3      & -    \\
        Oriented RCNN & 82.9 & 77.3 & 80.0      & 10.0  \\ \midrule
        DDQ-O (baseline) & 85.1 & 73.8 & 79.0 & 14.1 \\
        RQFormer    & 85.0 & 75.5 & \textbf{80.0}   & 14.1 \\ \bottomrule
        \end{tabular}
    }
    \label{ResultICDAR2015}
    \end{table}
}

\newcommand{\TableFPS}{
    \begin{table}[t]
    \centering
    \renewcommand\arraystretch{1.1}
    \caption{\textbf{Speed, Parameters and accuracy} on DOTA-v1.0.}
    \resizebox{\linewidth}{!}
    {
        \begin{tabular}{l|c|c|ccc}
        \toprule
        Method           & Frame & Backbone & FPS & Params (M) & mAP   \\ \midrule
        RoI Transformer& \multirow{3}{*}{\begin{tabular}[c]{@{}c@{}}Two-\\ stage\end{tabular}} &R50  &9.18 & 55.35  & 74.61 \\
        Oriented RCNN     & &R50   &7.30 & 41.37  & 75.87 \\
        Gliding Vertex  & &R101  &10.23& 41.36  & 75.02 \\ \midrule
        R3Det           & \multirow{5}{*}{\begin{tabular}[c]{@{}c@{}}One-\\ stage\end{tabular}} &R101 &6.12 & 42.13  & 73.79 \\
        CFA               & &R50  &16.59& 36.83  & 74.51 \\
        SASM             & &R50  &15.87& 36.83  & 74.92 \\
        PSC               & &R50  &16.81& 36.77  & 71.83 \\
        DCFL             & &R50  &17.73& 36.10  & 74.26 \\ \midrule
        D.DETR-O  & \multirow{3}{*}{\begin{tabular}[c]{@{}c@{}}End-\\ to-\\ end\end{tabular}}&R50  &10.83& 40.82  & 63.42 \\
        D.DETR-O w/CSL       & &R50 &10.71& 41.15  & 72.15 \\
        ARS-DETR         & &R50 &10.12& 41.56  & 73.79 \\ \midrule
        DDQ-O (baseline) & \multirow{2}{*}{\begin{tabular}[c]{@{}c@{}}End-\\ to-end\end{tabular}}&R50 &11.27& 63.71  & 73.57 \\
        RQFormer       &  &R50  &11.20& 40.90  & 75.04 \\ \bottomrule
        \end{tabular}
    }
    \label{FPS_Params}
    \end{table}
}

\newcommand{\TableIndividualDIORandDOTA}{
    \begin{table}[t]
    \centering
    \huge
    \renewcommand\arraystretch{1.3}
    \caption{Effectiveness of \textbf{individual modules} in RQFormer. DIOR-R and DOTA-v1.0 are used in this experiment.}
    \resizebox{\linewidth}{!}
    {
        \begin{tabular}{c|cc|ccc|ccc}
        \toprule
        \multirow{2}{*}{Method} & \multirow{2}{*}{RRoI Att.} & \multirow{2}{*}{SDQ} & \multicolumn{3}{c|}{DIOR-R} & \multicolumn{3}{c}{DOTA-v1.0} \\ \cmidrule{4-9}
                                &                            &                      & mAP     & Params(M)  & FPS  & mAP     & Params(M)   & FPS   \\ \midrule
        baseline                   &               &                      & 66.51   & 63.73    & 14.02 & 73.57  &  63.71        & 11.27\\
        RQFormer              &  \checkmark    &                      & 67.11   &  40.93    &14.42 & 74.05 &  40.90     & 11.82  \\
        RQFormer              &               &  \checkmark  & 67.27 &   63.73   & 13.54&74.89     &63.71    & 11.08  \\
        RQFormer              &   \checkmark   &   \checkmark  & 67.31   & 40.93   &14.00 & 75.04   &  40.90 &11.20  \\ \bottomrule
        \end{tabular}
    }
    \label{idividual_on_dior_dota1.0}
    \end{table}
}

\newcommand{\TableAblationStudiesIntegrated}{
    \begin{table*}[!htbp]
  \centering
    \caption{\textbf{RQFormer ablation experiments} with ResNet-50 on DIOR-R. The best result in each group is colored gray. 
    \vspace{1mm}
  }
  \hspace{-0.2cm}
  \subfloat[\small
  Effect of \textbf{pooling size} of RRoI Attention. Gradually increasing the pooling size, the performance is the best when it is 7.
  \label{ablation_pooling_size}
  ]{
     \begin{minipage}{0.3\linewidth}{
      \tablestyle{1.8pt}{1.3}
      \Tablepoolingsize
      }\end{minipage}
  }
        \hspace{0.5mm}
    \subfloat[\small
  Effect of \textbf{heads} of RRoI Attention. Gradually increasing the number of heads, the performance is saturated when it is 8.
  \label{ablation_attn_heads}
  ]{
      \begin{minipage}{0.3\linewidth}{
      \tablestyle{1.8pt}{1.3}
      \TableAttentionHead
      }\end{minipage}
  }
        \hspace{1em}
  \subfloat[\small
    \textbf{\footnotesize{RRoI attention} vs. other cross-attention.} Our RRoI attention outperforms dynamic attention and instance interaction.
  \label{ablation_attn_vs_others}
  ]{
      \begin{minipage}{0.3\linewidth}{
      \tablestyle{1.8pt}{1.2}
      \TableAttentionMechanism
      }\end{minipage}
  }
\\
    \subfloat[\small
    Effect of \textbf{IoU threshold} of SDQ. Increasing the threshold, the best performance occurs when it is 0.85.
  \label{ablation_iou_threshold}
  ]{
      \begin{minipage}{0.28\linewidth}{
      \tablestyle{1.8pt}{1.2}
      \TableIOUthreshold
      }\end{minipage}
  }
    \hspace{1em}
  \subfloat[\small
  \textbf{SDQ vs. other query selection} methods. Our proposed SDQ outperforms stack and top-$n$ method.
  \label{ablation_sdq_vs_others}
  ]{
      \begin{minipage}{0.31\linewidth}{
      \tablestyle{1.8pt}{1.2}
      \TableQuerySelectionMechanism
      }\end{minipage}
  }
  \hspace{1em}
  \subfloat[\small
  Effects of number of \textbf{decoder layers}. Gradually increasing the number of layers, the performance is saturated at 2 layers.
  \label{ablation_decoder_layer}
  ]{
      \begin{minipage}{0.31\linewidth}{
      \tablestyle{1.8pt}{1.2}
      \TableReduceDecoderLayer
      }\end{minipage}
  }
\\
 \hspace{-0.3cm}
  \subfloat[\small
    Different \textbf{combinations} of selective queries and distinct queries in the SDQ.
  \label{ablation_combination_of_sdq}
  ]{
       \begin{minipage}{0.45\linewidth}{
      \tablestyle{5pt}{1.3}
      \TabledecoderlayerinSDQ
       }\end{minipage}
  }
    \hspace{0.5cm}
  \subfloat[\small
  Ablation study on \textbf{each components} in RQFormer. Starting from DDQ-O, we gradually add RRoI attention and SDQ.
  \label{ablation_individual}
  ]{
       \begin{minipage}{0.45\linewidth}{
      \tablestyle{5pt}{1.2}
        \TableIndividualStrategy
        }\end{minipage}
  }   
\\
  \label{ablations}
  \end{table*}
}

\newcommand{\TableQueryNumber}{
    \begin{table}[!htbp]
    \renewcommand\arraystretch{1.1}
    \caption{Effect of positional and content \textbf{queries} on DIOR-R. Gradually increasing the number of queries, the performance is the best when it is 500.}
    \resizebox{\linewidth}{!}
    {    
        \begin{tabular}{c|ccccccc}
        \toprule
        Query  & 100 & 200 & 300 & 400 & \textbf{500} & 600 & 700 \\
        \midrule
        mAP    & 60.23 & 61.57 & 62.76 & 65.36 & \textbf{67.31} & 66.84 & 66.51 \\ 
        Params & 40.8M & 40.8M & 40.9M & 40.9M & \textbf{40.9M} & 40.9M & 41.0M \\
        \bottomrule
        \end{tabular}
    }
    \label{query_number}
    \end{table}
} 

\begin{document}
\begin{sloppypar}

\let\WriteBookmarks\relax
\def\floatpagepagefraction{1}
\def\textpagefraction{.001}
\shorttitle{RQFormer: Rotated Query Transformer for End-to-End Oriented Object Detection}
\shortauthors{Zeyu Ding et~al.}

\title [mode = title]{RQFormer: Rotated Query Transformer for End-to-End Oriented Object Detection}

\author[1,2]{\textcolor{black}{Zeyu Ding}}[style=chinese]
\ead{dingzeyu@cumt.edu.cn}
    
\author[1,2]{\textcolor{black}{Yong Zhou}}[style=chinese]
\ead{yzhou@cumt.edu.cn}
\cormark[1]

\author[1,2,3]{\textcolor{black}{Jiaqi Zhao}}[style=chinese]
\ead{jiaqizhao@cumt.edu.cn}
    
\author[1,2]{\textcolor{black}{Hancheng Zhu}}[style=chinese]
\ead{zhuhancheng20@cumt.edu.cn}
    
\author[1,2]{\textcolor{black}{Wen-Liang Du}}[style=chinese]
\ead{wldu@cumt.edu.cn}
    
\author[1,2]{\textcolor{black}{Rui Yao}}[style=chinese]
\ead{ruiyao@cumt.edu.cn}
    
\author[4]{\textcolor{black}{Abdulmotaleb El Saddik}}
\ead{elsaddik@uottawa.ca}

\address[1]{School of Computer Science and Technology, China University of Mining and Technology, Xuzhou 221116, China}
\address[2]{Mine Digitization Engineering Research Center of the Ministry of Education, Xuzhou 221116, China}
\address[3]{Innovation Research Center of Disaster Intelligent Prevention and Emergency Rescue, Xuzhou 221116, China}
\address[4]{School of Electrical Engineering and Computer Science, University of Ottawa, Ottawa, ON K1N 6N5, Canada}

\cortext[cor1]{Corresponding author}

\begin{abstract}
Oriented object detection presents a challenging task due to the presence of object instances with multiple orientations, varying scales, and dense distributions. Recently, end-to-end detectors have made significant strides by employing attention mechanisms and refining a fixed number of queries through consecutive decoder layers. However, existing end-to-end oriented object detectors still face two primary challenges: 1) misalignment between positional queries and keys, leading to inconsistency between classification and localization; and 2) the presence of a large number of similar queries, which complicates one-to-one label assignments and optimization. To address these limitations, we propose an end-to-end oriented detector called the Rotated Query Transformer, which integrates two key technologies: Rotated RoI Attention (RRoI Attention) and Selective Distinct Queries (SDQ). First, RRoI Attention aligns positional queries and keys from oriented regions of interest through cross-attention. Second, SDQ collects queries from intermediate decoder layers and filters out similar ones to generate distinct queries, thereby facilitating the optimization of one-to-one label assignments. Finally, extensive experiments conducted on four remote sensing datasets and one scene text dataset demonstrate the effectiveness of our method. To further validate its generalization capability, we also extend our approach to horizontal object detection.
\end{abstract}




\maketitle

\section{Introduction}

Oriented object detection~\citep{WEN2023119960} aims to predict categories and a set of oriented bounding boxes for each target, and it is widely used in remote sensing~\citep{infofpn,FDLR-Net,clt-det} and scene text detection. Unlike horizontal object detection, which focuses on natural scenes, oriented object detection remains a challenging task due to the diversity of oriented objects, such as arbitrary orientations, varying scales, and dense distributions. Although modern oriented detection methods, particularly those based on convolutional neural networks (CNNs), have achieved excellent results, challenges persist. These methods can be categorized into two types: single-stage~\citep{dcfl,psc} and two-stage~\citep{orientedrcnn,redet} approaches. Single-stage methods benefit from numerous anchors (position priors) and a straightforward architecture, while two-stage methods are inherently designed to learn rich features through a coarse-to-fine training style.

Inspired by the attention and transformer in natural language processing (NLP)~\citep{attention_is_all_you_need}, many detectors introduce it into horizontal object detection to extract global features and show promising performance~\citep{detr,deformabledetr}. Most end-to-end horizontal object detectors rely on an encoder-decoder framework and remove complicated hand-designed modules.

However, when the end-to-end framework is integrated into oriented object detection, it will suffer from several challenges. First, \textbf{\textit{objects appear in arbitrary orientations leading to misalignment between positional queries and keys}}, shown in Fig. \ref{heat_map}(a). Image features as values and oriented boxes as positional queries need to be aligned when they are interacted with in cross-attention. In addition, the scales of objects vary. Small and large objects require high and low resolution features respectively. Therefore values are extracted from multiple levels of features to adequately represent objects. Second, \textbf{\textit{similar queries hinder optimization.}} The end-to-end detectors adopt one-to-one label assignments. Consider the situation where two queries are entirely identical. When label assignment, one query is assigned as a positive sample, while the other is assigned as a background category. Such a problem can potentially introduce adverse effects into the training process~\citep{ddq}. To alleviate it, some horizontal detectors~\citep{dino, hdetr} have increased the number of queries by more than 1000. However, it's common to find hundreds of densely packed object instances within a single image in oriented object detection. Dense objects require a large number of queries. Applying the same rate of increase to dense object queries will result in a significant increase in computational overhead.

\heatmap

In this paper, we propose an end-to-end oriented object detector to address the above problems called Rotated Query Transformer (RQFormer), which consists of two effective technologies, Rotated RoI attention (RRoI attention) and Selective Distinct Queries (SDQ). Specifically, the proposed RRoI attention can align keys with positional queries through an attention mechanism. Learnable oriented boxes as positional queries extract aligned features that serve as the values in this attention mechanism. Content queries generate attention weights and aggregate the values naturally. The features around the center point of positional queries and within them are highlighted and scored by the attention mechanism, shown in Fig. \ref{heat_map}(b). The time complexity of RRoI Attention is of a similar magnitude to deformable attention~\citep{deformabledetr}.

In addition, our proposed Selective Distinct Queries (SDQ) collects queries from intermediate decoder layers once and then filters similar queries to obtain distinct queries. In modern end-to-end detectors, the decoder consists of 6 consecutive decoder layers~\citep{ddq, sparsercnn, deformabledetr} and each decoder layer predicts results supervised by loss functions. We have observed that queries in the intermediate layers can represent objects adequately compared to initial queries. Hence, we collect queries from intermediate layers as high-quality selective queries, without introducing a large number of initial queries~\citep{dino} or extra auxiliary network branches~\citep{hdetr,co_detr}. These queries can effectively cover objects. But there are still a large number of similar queries~\citep{ddq} within the query set. These similar queries hinder the optimization of training. Our method can significantly filter similar queries and keep the remaining queries distinct, which facilitates one-to-one label assignment and training. The number of distinct queries is dynamic and normally less than the initial ones. SDQ is embedded into intermediate decoder layers, rather than positioned before or after the entire decoder.

Our contribution can be summarized in three-folds:

\begin{itemize}
\item To address the misalignment, the Rotated RoI attention is proposed to align positional queries and keys via a cross-attention mechanism for objects with arbitrary orientations and varying scales towards oriented object detection tasks.
\item To mitigate the issue of query similarity, the proposed Selective Distinct Queries first collects sparse high-quality queries from the intermediate decoder layers. Second, it filters similar queries to obtain distinct queries, which can facilitate one-to-one label assignment and training.
\item Extensive experiments on six datasets demonstrate the effectiveness of our approach, including oriented object detection, text detection, and horizontal object detection.
\end{itemize}

\section{Related Work}

\subsection{End-to-end Oriented Object Detection}

End-to-end transformer-based methods~\citep{detr, deformabledetr, dino} have demonstrated promising performance in horizontal object detection tasks. Inspired by them, some researchers have extended Transformer-based approaches to oriented object detection. AO2-DETR~\citep{ao2detr} adapt Deformable DETR~\citep{deformabledetr} to oriented object detection task. It designs oriented box generation and refinement modules to provide accurate position prior. ARS-DETR~\citep{arsdetr} specializes in adapting DN-DETR~\citep{dn_detr} for the oriented object detection task. It proposes the rotated deformable attention, where sampling points are rotated based on angles for feature alignment. RHINO~\citep{rhino} takes a route by adapting DINO~\citep{dino} for oriented object detection and improves L1 distance cost to Hausdorff distance cost in Hungarian Matching of one-to-one label assignment for more stable training. In addition, several studies have focused on enhancing one-to-one label assignment. EMO2-DETR~\citep{emo2detr} observes and addresses the issue that one-to-one label assignment results in relative redundancy of object queries because objects are unevenly distributed in images. Furthermore, certain methods focus on improving object queries. PSD-SQ~\citep{psd-sq} represents object queries as point sets rather than oriented boxes for accurate instance feature sampling. D$^2$Q-DETR~\citep{d2qdetr} designs dynamic queries which gradually reduce the number of object queries in the stacked decoder layers to better balance model precision and efficiency. Different from the above methods, our proposed approach introduces Rotated RoI attention for alignment between positional queries and keys and Selective Distinct Queries to facilitate one-to-one label assignment.

\subsection{Attention in Object Detection}
DETR~\citep{detr} first introduces Transformer~\citep{attention_is_all_you_need} into object detection, but the single feature map and vanilla attention are low efficiency to scale and large resolution images. The spatially modulated co-attention proposed in SMCA~\citep{smca} combines vanilla attention with a Gaussian-like weight map to constrain co-attention to focus more on regions near the predicted box. The deformable attention modules proposed by Deformable DETR~\citep{deformabledetr} only focus on some key sampling points around a reference. These key sampling points are responsible for both classification and regression, but at the supervision of orientation, they are located at special positions that are sub-optimal, like the catercorner and axis of boxes. Anchor DETR~\citep{anchor_detr} designs Row-Column Decoupled Attention that performs the row attention and column attention successively. The decoupled row and column attention lose spatial and orientation information. Some works~\citep{dynamic_detr} realize attention using dynamic convolution. Different from the above attentions, our proposed Rotated RoI attention (RRoI attention) focuses on interested regions of positional queries $B$ ($cx,cy,w,h,\theta$) distributed in multi-scale features and aligns keys to oriented objects. The time complexity of RRoI Attention is the same order of magnitude as deformable attention.

\overallframework

\subsection{Methods of Query Update}
We classified current methods of query update into three types: the basic type, query enhancement type, and auxiliary decoder branch type. In all these types, queries are successively updated layer by layer in the decoder. In the basic way, a fixed number of queries which are initialized randomly are fed into the first decoder layer and the outputs are also the same number of initial queries. Subsequently, these queries are decoded successively by other decoder layers, each sharing the same structure~\citep{detr,sparsercnn,deformabledetr}. Due to the sparsity and random initialization of the queries, some enhancement methods are proposed. Anchor DETR~\citep{anchor_detr} attaches three patterns to each anchor point which enlarges query numbers. Dynamic Sparse R-CNN~\citep{dynamic_sparse_rcnn} replaces random initialization with a dynamic proposal generation mechanism to initialize proposals and features. DDQ~\citep{ddq} supervises dense boxes lying on features and selects top-scoring distinct proposals as inputs to the decoder. SQR~\citep{sqr} stacks queries from previous decoder layers and feeds them to later layers. DQ-Det~\citep{dq_det} combines basic queries as modulated queries according to the convex combination. Some researchers introduce an auxiliary decoder branch to accelerate training convergence. To realize one-to-may label assignment, Group DETR~\citep{group_detr} feeds multiple groups of queries to decoders as auxiliary branches in parallel, similar to the way in H-DETR~\citep{hdetr}. DN-DETR~\citep{dn_detr} and DINO~\citep{dino} add noises to bounding boxes and class labels as noising queries in an auxiliary branch to mitigate the instability of bipartite graph matching. Co-DETR~\citep{co_detr} train multiple auxiliary heads supervised by one-to-many label assignments. Different from the above query learned way, our Selective Distinct Queries (SDQ) collects sparse high-quality queries from intermediate decoder layers which avoid a large number of initial queries or auxiliary decoder branches. To mitigate the issue of query similarity, SDQ filters similar queries and obtains distinct queries. Moreover, SDQ is embedded among decoder layers, rather than served as an auxiliary head. Without losing generality, SDQ can be easily combined with other enhancement methods.

\section{PROPOSED METHOD}
In this paper, we propose an end-to-end oriented detector called RQFormer which is equipped with the Rotated RoI attention (RRoI attention) and Selective Distinct Queries (SDQ). We first introduce the overall architecture in Sec. \ref{overall_architecture}. Specifically, we illustrate the RRoI attention and SDQ in Sec. \ref{RRoI_attention} and \ref{Selective_distinct_queries}, respectively. Last, we introduce losses in Sec.~\ref{label_assignment_and_losses}.

\RRoIAttn

\subsection{Overall Architecture}
\label{overall_architecture}

Our RQFormer consists of a backbone, a Feature Pyramid Network (FPN), multiple decoder layers with Rotated RoI attention, the Selective Distinct Queries (SDQ) module, label assignment, and losses, shown in Fig. \ref{overall_framework}.

\textbf{Backbone and FPN}. The inputs of the backbone are images $x_{img}\in \mathbb{R}^{3\times H_0 \times W_0}$, where 3, $H_0$, and $W_0$ denote color channels, heights, and widths of original images respectively. The output multi-scale features $\{x^l\}_{l=1}^{L}$ are extracted by backbone (e.g., ResNet50~\citep{resnet}) and then FPN, where $x^l\in \mathbb{R}^{C\times H_{l}\times  W_{l}}$, $L$ denotes levels of features, and $C$ represents channels of features. The scales of objects in oriented object detection vary. Small objects need high-resolution features, while large ones benefit from low-resolution features. Therefore, we use multi-scale features instead of a single feature level.

The vast size of images in remote sensing scenes results in high-resolution image features. To mitigate the computational burden, we do not introduce a dedicated image encoder for further interaction with features, in line with~\citep{rtdetr, litedetr}. Only the FPN is used to feature fusion.

\textbf{Content Queries and Positional Queries.} Learnable positional and content queries serve as inputs to the decoder for representing object instances. The learnable positional queries $B\in \mathbb{R}^{N\times 5}$ denote coordinates of objects as $(cx,cy,w,h,\theta)$, signifying the center coordinate, width, height, and angle, respectively. Content queries $Q\in \mathbb{R}^{N\times C}$ are learnable high-dimension latent vectors, capturing semantic information of objects, where $C$ denotes dimensions of content queries, numerically the same as the channel of image features. Our method initializes positional and content queries through DDQ~\citep{ddq}. They undergo sequential updates within the decoder, layer by layer. The number $N$ of content queries always equals that of positional queries. Learnable positional queries handle regression, while content queries manage classification.

\textbf{Self-attention in Decoder.} Self-attention enables the model to inhibit duplicate predictions~\citep{detr} and facilitates interaction between content queries. The inputs of self-attention are content queries $Q$ and positional encoding $PE\in \mathbb{R}^{N\times C}$. The positional encoding comprises learnable embeddings~\citep{detr}, initialized with the same shape as content queries. The self-attention is calculated as follows:
\begin{equation}
    \begin{split}
    {\rm MHSA}(Q,K_q,V_q)=\\{\rm softmax}((Q&+PE)(K_q+PE)^\top/\sqrt{d_q})V_q
    \end{split}
\label{eq_MHSA}
\end{equation}
where $K_q$ and $V_q$ are generated from content queries, sharing the same dimensions. The $d_q$ denotes dimensions. The outputs of self-attention keep the same shapes as the inputs.

\textbf{Cross-attention and SDQ in Decoder.} Our decoder incorporates two technologies, Rotated RoI attention (RRoI attention) and Selective Distinct Queries (SDQ). RRoI attention serves as cross-attention, aligning positional queries with keys through an attention mechanism.  The decoder comprises consecutive layers, with SDQ embedded between two intermediate layers. SDQ can avoid a large number of initial queries and facilitate one-to-one label assignment and training. In the next part, we will provide a detailed elaborate on these components.

\subsection{Rotated RoI Attention}
\label{RRoI_attention}
\textbf{Decoder with Rotated RoI Attention.} Both cross-attention and self-attention modules are in the decoder, shown in Fig. \ref{fig_2}(a). A small set of learnable content queries $Q\in \mathbb{R}^{N\times C}$  are high-dimension (e.g., $C=256$) latent vectors and are expected to encode semantic information of objects. The positional encoding embeddings are randomly initialized with the same shape as queries and then added by queries following~\citep{detr}. The content queries and positional encodings are jointly trained with the networks. In self-attention modules, content queries interact with each other and the outputs keep the same shapes as the inputs. In our proposed Rotated RoI attention, as cross-attention, content queries interact with corresponding regions of aligned features. The inputs of RRoI attention include content queries from self-attention, positional queries, and multi-scale feature maps. Though the content queries represent object characteristics, we introduce positional queries $B\in \mathbb{R}^{N\times 5}$ to localize the objects. These positional queries are represented by parameters ($cx,cy,w,h,\theta$), denoting the center coordinate, width, height, and angle, respectively. The number of positional queries always equals content queries. Both content and positional queries are initialized through DDQ~\citep{ddq}. The multi-scale feature maps $\{x^l\}_{l=1}^{L} $, where $x^l\in \mathbb{R}^{C\times H_{l}\times  W_{l}}$, are extracted by backbone (e.g., ResNet50~\citep{resnet}) and Feature Pyramid Network (FPN). In the next part, we will discuss RRoI attention modules specifically.

\textbf{Multi-Scale Rotated RoI Attention Module.}
Given positional queries $B\in \mathbb{R}^{N\times 5}$ and multi-scale feature maps $\{x^l\}_{l=1}^{L} $ in Fig. \ref{fig_2}(b), to solve the misalignment problems, we first align the features as values $V$,
\begin{equation}
    V={\rm Concat}\{\{R_{l}(x^{l},\phi (B_{n}),r) \}_{n=1}^{N}\}_{l=1}^L
  \label{eq_1}
\end{equation}
where integer $l$ index the input feature level, and $n$ index the positional queries. The values $V\in \mathbb{R}^{N\times C \times r^{2}}$ will further interact with content queries. Function $\phi (B_{n} )$ maps positional queries to different feature levels by their scales. Function $R_{l}(\cdot )$ warps region features from the corresponding feature $x^l$ precisely according to positional queries and pooling size $r$, e.g., Rotate RoI Align~\citep{roi_transformer}. The function $R_{l}(\cdot)$ is calculated as follows:
\begin{equation}
\begin{split}
  &R_l(x^l,\phi(B_n),r)=\mathrm{Concat}\{\{\mathrm{Bi}(x^l[\omega(i,j)])\}_{i=1}^r\}_{j=1}^r,\\
  & \omega (i,j)=
    \begin{pmatrix}
     \cos \theta_n & \hspace{-0.5em} -\sin \theta_n \\
     \sin \theta_n & \hspace{-0.5em} \cos \theta_n
    \end{pmatrix} 
    \begin{pmatrix}
     (cx_n-\frac{w_n}{2}+\frac{iw_n}{r})\frac{W_l}{W_0}\\
     (cy_n-\frac{h_n}{2}+\frac{jh_n}{r})\frac{H_l}{H_0}
    \end{pmatrix}
\end{split}
\end{equation}
The focused values $V$ are calculated using bilinear operator $\mathrm{Bi}(\cdot)$ on the corresponding feature $x^l$. The indices of the focused regions are calculated by $\omega(i,j)$. The concatenation is used because the images and features lie on a 2D plane, taking into account both the horizontal and vertical axis.

The attention weight $A$ is obtained via linear projection over the queries. The learnable attention weights help RRoI attention focus on different parts of the values and heads.
\begin{equation}
    A=softmax(Linear(Q))
\end{equation}
Specifically, the content queries $Q$ are first fed to a linear projection operator of $M \times r^2$ channels and then are fed to a softmax operator to obtain the attention weight $A \in \mathbb{R}^{N \times M \times r^2}$. The $M$ denotes the number of attention heads.

Given content queries $Q\in \mathbb{R}^{N\times C}$, the multi-scale Rotated RoI Attention is calculate by
\begin{equation}
\begin{split}
        {\rm MSRRoIAttn}(Q_{n}, B_{n},\{x^l\}_{l=1}^{L},r)=\\
        \sum_{m=1}^{M}W_{m}[\sum_{k=1}^{r^2}A_{nmk}\cdot(W_{m}'V)_{nmk} ]
\end{split}
  \label{eq_2}
\end{equation}
where $n$ indexes the positional and content queries, $m$ indexes the attention heads and  $k$ indexes all $r^2$ points at a single value map. $A_{nmk}$ denotes the attention weight of the $k^{th}$ point at a single value map in the $m^{th}$ head. The scalar attention weight $A_{nmk}$ is normalized by $\sum_{k=1}^{r^2} A_{nmk}=1$. The $W_{m}'$ and $W_{m}$ are input value projection and output projection matrix at $m^{th}$ head respectively.

\textbf{Complexity For Rotated Rotated RoI Attention.}
When calculating the values, to align features for positional queries, the bilinear interpolation will be used. Four adjacent pixel values interpolate one precise target point. So the complexity for calculating values is $O(4Nr^{2}s)$, where $s$ denotes the sampling ratio and $r^{2}s$ denotes all sampling points in one positional query. Then for projecting values to multiple heads, the complexity is $O(r^{2}NC^{2})$. Due to attention weights obtained via linear projection over queries, the complexity is $O(MNCr^2)$. When attention weights and values interact, it is $O(NCr^2)$. Finally, for aggregating sampled values, the complexity is $O(NC^{2})$. So the overall complexity of Rotated RoI attention is $O(4Nr^{2}s+r^{2}NC^{2}+NMCr^{2}+NCr^{2}+NC^{2})$. In our experiments, $M=8$, $r=7$, $s=4$ and $C=256$ by default, thus $r^{2}<C$ and the complexity is of $O(N(50C^{2}+441C+784))$. The complexity is of a similar magnitude to deformable attention.

\SDQ

\subsection{Selective Distinct Queries}
\label{Selective_distinct_queries}

\textbf{Basic Query Update in Decoder.} Before the first decoder layer, the content and positional queries are initialized as inputs to the decoder, shown in Fig. \ref{fig_3}(a). For brevity, both initial content and positional queries are denoted by $q^{0}=\{q_{1}^{0},q _{2}^{0},\cdot \cdot \cdot ,q_{n}^{0} \}$ here. In the basic query update mode, there are $d$ identical decoder layers (e.g., $d=6$) stacked successively. Each decoder layer $D^{i}$ takes the queries $q^{i-1}$ from the previous layer and processes them to obtain the updated queries $q^{i}$ for the next layer,
\begin{equation}
    q^{i}=D^{i}(q^{i-1})
  \label{eq_3}
\end{equation}
where $i=1,2,\cdot\cdot\cdot,d$ and $d$ is the total number of decoder layers. After passing through all decoder layers, the model would typically have generated refined queries $q^{d}$. The loss $\mathcal{L}=\lambda_{cls}\mathcal{L}_{cls}+\lambda_{L_{1}}\mathcal{L}_{1}+\lambda_{iou}\mathcal{L}_{iou}    $ is applied at the queries of all the decoder layers for training, where $\mathcal{L}_{cls}$ is the focal loss~\citep{focal_loss} for classification, $\mathcal{L}_{1}$ and $\mathcal{L}_{iou}$ are L1 Loss and Rotated IoU~\citep{rotated_iou_loss} for regression.

\Tableone

\textbf{Selective Queries.}
As shown in Table \ref{table_1}, it appears that there is a clear growing trend in mAP as queries move deeper into the decoder layers. Notably, the mAP can be easily tested by adjusting the number of decoder layers, since the outputs of each decoder layer are supervised by loss functions. Taking ARS-DETR~\citep{arsdetr} as an example, the mAP is 33.07\% and 39.63\% at the $4^{th}$ and $5^{th}$ decoder layer respectively. This suggests that the queries in the last few decoder layers already capture sufficient information about object positions and categories. Therefore, we can collect the queries from $p$ intermediate decoder layers as selective queries only once, and similar queries among them will be filtered to obtain distinct queries in the next step, shown in Fig. \ref{fig_3}(b).
\begin{equation}
    q^{p}={\rm Concat}(q^{j},q^{j-1},\cdot \cdot \cdot ,q^{j-p-1})
  \label{eq_4}
\end{equation}
where $q^{p}$ denotes as selective queries from the $j^{th},(j-1)^{th},\cdot \cdot \cdot ,(j-p-1)^{th}$ intermediate decoder layers, totally $p$ layers. The number of selective queries is $p\times   n$, where $n$ is the initial number of queries. Different $p$ intermediate decoder layers collect $p$ times initial queries. These selective queries avoid a large number of initial queries and multiple auxiliary branches.

\textbf{Distinct Queries.}
The selective queries collected from intermediate decoder layers already capture sufficient information about the positions and categories with the images. But there exist many similar queries. To obtain distinct queries, similar queries will be filtered according to:
\begin{equation}
\hat{q}  = \begin{cases}
\text{top-} n(\psi (q^{p} )),\quad &|\psi (q^{p} )|>n \\
\psi (q^{p} ),\quad &|\psi (q^{p} )|\le n
\end{cases}
  \label{eq_5}
\end{equation}
where $\hat{q}$ denotes distinct queries. Distinct queries are calculated in two steps. First, we filter similar queries by a simple class-agnostic non-maximum suppression (NMS), denoted as $\psi(\cdot )$. Second, if the remaining queries $\psi(q^{p} )$ are more than the initial queries $n$, we will choose the first $n$ ones according to scores as distinct queries. Otherwise, all remaining queries are treated as distinct queries. The class-agnostic NMS operation here is regarded as the first step of distinct queries mechanisms embedded among intermediate decoder layers, instead of as post-processing at the inference stage in traditional detectors~\citep{orientedrcnn,redet}. Therefore, such a method still conforms to the end-to-end standard. In the basic queries update way, the last decoder layer $D^{d}$ exclusively refines queries from the $D^{d-1}$, while our method handles queries from multiple intermediate layers simultaneously. Through filtering similar queries, the number of distinct queries is reduced which reduces computational overhead.

\whysdqwork

\textbf{Analysis of Selective Distinct Queries.}
Our proposed SDQ can filter similar queries and keep the remaining queries distinct. The one-to-one label assignment was first introduced to object detection by DETR~\citep{detr}. Only one sample can be assigned to a corresponding ground truth object according to matching cost matrices, while all other samples are assigned to the background class. However, there exists a large number of similar queries which hinder training optimization. We use Intersection over Union (IoU) to measure the similarity of queries at the stage of one-to-one label assignment. If the IoU between a positive sample and a background query is greater than the threshold, they are regarded as similar. We define the ratios of similar queries as:
\begin{equation}
   \mathrm{ratios}=\frac{\sum_{i=1}^{N_{\mathrm{pos}}}\mathbbm{1}(\max(\mathrm{IoU}(\mathrm{pos}_{i},\mathrm{neg}_{j}))>t)}{\sum_{i=1}^{N_{\mathrm{GT}}}\mathrm{GT}_{i}}
  \label{eq_sdq}
\end{equation}
where $\mathbbm{1}(\cdot)$, $\mathrm{pos}$, $\mathrm{neg}$, $t$, $\mathrm{GT}$ denote indicator function, positive samples, negative samples, threshold, and ground truth objects. Note that the number of positive samples $N_{\mathrm{pos}}$ and ground truth objects $N_{\mathrm{GT}}$ is always equal. Why only use IoU to measure the similarity of queries, rather than considering all matching cost matrices including IoU, L1 distance, and Focal loss? There are two main reasons. First, the model adopts the one-to-one label assignment. Second, there are no ground truth objects with an IoU greater than a threshold close to 1, e.g., 0\% ground truth objects with an IoU greater than 0.8 in the DIOR-R test set~\citep{dior}. To quantify the similarity among queries, we conduct an analytical experiment on the DIOR-R test set shown in Fig. \ref{why_sdq_work}. Taking $t=0.95$ as an example, 4.00\% of queries showed similarity in baseline model DDQ-O~\citep{ddq}. Our SDQ can significantly reduce the value to 0.74\% which can prove the effectiveness of our method.

\subsection{Prediction, Label Assignment and Loss}
\label{label_assignment_and_losses}

In each decoder layer, the model predicts categories and boxes. This prediction is achieved through a simple Feedforward Neural Network (FFN) and linear layers.

Object detection tackles two sub-tasks, regression for positions and classification for categories. To execute these tasks, bipartite matching is applied at the label assignment stage between predictions and ground truth objects. The matching cost is calculated as follows:
\begin{equation}
\mathcal{L}=\lambda_{iou}\cdot\mathcal{L}_{iou}+\lambda_{L_1}\cdot\mathcal{L}_{L_1}+\lambda_{cls}\cdot\mathcal{L}_{cls}
  \label{eq_losses}
\end{equation}
where $\mathcal{L}_{iou}$ and $\mathcal{L}_{L_1}$ are Rotate IoU loss~\citep{rotated_iou_loss} and $L_1$ loss between predicted boxes and ground truth boxes, respectively. $\mathcal{L}_{cls}$ is focal loss~\citep{focal_loss} for classification between predicted categories and ground truth labels. $\lambda_{iou}$, $\lambda_{L_1}$ and $\lambda_{cls}$ are coefficient of corresponding component. Based on matching cost $\mathcal{L}$, only one optimal sample is assigned to each ground truth object, while the remaining samples are assigned to the background.

Training losses are mathematically the same as matching costs, but they are performed on the matched pairs. The losses are applied to all predicted content and positional queries across all decoder layers.

\section{EXPERIMENTS}
\subsection{Datasets}

We conduct our experiments on six public datasets, including oriented object detection datasets DIOR-R~\citep{dior}, DOTA-v1.0~\citep{dotav1.0}/v1.5/v2.0~\citep{dota-v2.0},  a scene text detection dataset ICDAR2015~\citep{icdar2015}, and a horizontal object detection dataset MS COCO~\citep{coco}.

\textbf{DIOR-R}~\citep{dior} is a large-scale oriented object detection dataset. It contains a total of 23,463 remote sensing images and 192,512 instances covering 20 common classes including windmill (WM), trainstation (TS), storagetank (STO), ship (SH), harbor (HA), golffield (GF), expressway toll station (ETS), bridge (BR), baseballfield (BF), airplane (APL), vehicle (VE), tenniscourt (TC), stadium (STA), overpass (OP), groundtrackfield (GTF), dam (DAM), expressway service area (ESA), chimney (CH), basketballcourt (BC) and airport (APO). We train our model on training and validation sets and report the results on the test set.

\textbf{DOTA} series including DOTA-v1.0/v1.5/v2.0 are used. \textbf{DOTA-v1.0}~\citep{dotav1.0} is designed for oriented object detection tasks which contain 188,282 instances and 2,806 remote sensing images ranging in size from 800$\times$800 to 4,000$\times$4,000. It covers 15 different object categories: Plane (PL), Baseball diamond (BD), Bridge (BR), Ground track field (GTF), Small vehicle (SV), Large vehicle (LV), Ship (SH), Tennis court (TC), Basketball court (BC), Storage tank (ST), Soccer-ball field (SBF), Roundabout (RA), Harbor (HA), Swimming pool (SP), and Helicopter (HC). \textbf{DOTA-v1.5} uses the same images as DOTA-v1.0 but adds a new category Container Crane (CC) and more small instances, which contains 403,318 instances. \textbf{DOTA-v2.0} adds two categories Airport and Helipad compared to DOTA-v1.5, which contains 11,268 images and 1,793,658 instances. We use both training and validation sets for training and test sets for testing for the DOTA series. The detection result of the test set is obtained by submitting testing results to DOTA’s official evaluation server.

\textbf{ICDAR2015}~\citep{icdar2015} is a text detection dataset, which includes 1,000 training images and 500 test images.

\textbf{MS COCO 2017}~\citep{coco} is a horizontal object detection dataset for natural scenes, consisting of 118K images as train2017 for training and 5K images as val2017 for validation. We train our model on train2017 and report the evaluation results on val2017.

\subsection{Implementation Details}
We select ResNet-50~\citep{resnet} as the backbone which is pre-trained on the ImageNet~\citep{imagenet}. We optimize the overall network with AdamW optimizer~\citep{adamw} with the warming up for 500 iterations. We use Focal loss~\citep{focal_loss} for classification and Rotated IoU/GIoU loss~\citep{rotated_iou_loss} and L1 loss for regression.

For experiments on the DOTA-v1.5/v2.0 and ICDAR2015, we use 2 NVIDIA RTX 2080ti with a batch size of 4 (2 images per GPU) and the learning rate of $5\times e^{-5}$ for training, while a single GPU with a batch size of 2 and the learning rate of $2.5\times e^{-5}$ on DOTA-v1.0 and DIOR-R. Models build on MMRotate~\citep{mmrotate} with Pytorch. The weights of losses are 2.0, 2.0, and 5.0 for Focal loss, L1 loss, and Rotated IoU loss respectively. Following the official settings of the DOTA benchmark, we crop images into patches of $1024\times 1024$ with overlaps of 200 and train the model for 24 epochs. Images on DIOR-R are trained for 36 epochs with the original fixed size of $800\times 800$, while 12 epochs in ablation studies. To compare with other methods, the number of queries is 500, while in ablation studies on DIOR-R, it is 300. Images on ICDAR2015 are trained for 160 epochs with a fixed size of $800 \times 800$.

For experiments on COCO 2017, we use 2 NVIDIA RTX 2080ti with a batch size of 4 (2 images per GPU) for training. Models build on MMDetection~\citep{mmrotate} with Pytorch. The weights of loss are 2.0, 5.0, and 2.0 for Focal loss, L1 loss, and IoU loss respectively. The learning rate is $2\times e^{-5}$.

\TableAblationStudiesIntegrated

\subsection{Ablation Study}
\textbf{Different pooling size.}
To further explore RRoI Attention, we adopt different pooling sizes $r$. As the pooling size $r$ increases, more features will be highlighted in positional queries. When pooling size $r$ is set to 5, the AP is only 61.35\% on DIOR-R, shown in Table \ref{ablations} (a). When it is increased to 7, the mAP can achieve 62.63\%. This suggests that abundant features are sufficient to represent objects. When we increase the pooling size to 9, the performance of the model declines. It indicates that redundant features confuse the attention mechanism. As the pooling size gradually increases, the model's parameters also increase, given that the parameters of values $V$ become larger.

\textbf{The number of attention head.}
Following deformable attention~\citep{deformabledetr}, our RRoI attention also adopts multiple heads. Different heads establish different associations between queries and values. In this ablation, we use different numbers of $m$ heads. As we increase the number of heads from 2 to 8, the mAP grows from 61.36\% to 62.63\% on DIOR-R, shown in Table \ref{ablations} (b). This indicates that the multi-head attention extends the ability to focus on different parts of features, and provides multiple subspaces for representation. When we increase the number of heads to 16, there is a decline in performance. We suspect that the redundant head focuses on redundant features. As the number of heads increases, the model's parameter count rises, driven by the increased number of parameters in values $V$, attention weights $A$, and projection matrices.

\textbf{The RRoI attention vs. other cross-attention.}
To further affirm the efficacy of RRoI attention, we conduct a comparative analysis with other attention mechanisms, namely instance interaction~\citep{sparsercnn} and dynamic attention~\citep{dynamic_detr}, shown in Table \ref{ablations} (c). Although instance interaction is not strictly an attention mechanism, we discuss it for brevity. RRoI attention attains an mAP of 62.76\%, outperforming instance interaction by 0.35\% and dynamic attention by 1.17\%. The model incorporating RRoI attention exhibits the lowest parameter count, standing at 40.88M, which is 22.8M less than instance interaction and 6.09M less than dynamic attention. This reduction is attributed to RRoI attention's focus on the region inside boxes, resulting in significant computational efficiency. Moreover, the model with RRoI attention achieves faster detection speed, reaching 14.28 FPS, surpassing both instance interaction and dynamic attention.

\textbf{Threshold of Selective Distinct Queries.}
To further explore the effectiveness of SDQ, we conducted ablation experiments on the different IoU thresholds shown in Table \ref{ablations} (d). To filter the similar queries and keep the remaining queries distinct, the threshold is a high value close to 1 in SDQ. As the IoU threshold value increases from 0.8 to 0.85, the mAP increases by 0.18\% on DIOR-R. When the IoU threshold increases to 0.95 which is very close to 1, the mAP value decreases slightly, indicating that there are very few predicted boxes that fully coincide or are close to perfectly coinciding. The number of parameters is the same at different thresholds.

\TableDIORResult
\TableDOTAVoneResult

\textbf{SDQ vs. other queries selection mechanism.}
To further verify the effectiveness of SDQ, we performed ablation experiments with different selection mechanisms on DIOR-R, including top-$n$ and stack mechanisms, shown in Table \ref{ablations} (e). Following DINO~\citep{dino}, we choose top-$n$ queries according to scores of multi-class classification, while in Deformable DETR~\citep{deformabledetr}, the top-$n$ method selects proposals only according to scores of binary classification. In the top-$n$ mechanism, we collect queries from $1^{st}$ and $2^{nd}$ decoder layers and choose top-$n$ queries at $2^{nd}$ layer. The number of $n$ is equal to initial queries. In addition, we densely stack queries from $1^{st}$ and $2^{nd}$ decoder layers to $3^{th}$ layers which can be regarded as a revision of SQR~\citep{sqr}. The number of queries multiplies many times according to candidate layers through the stack. Essentially, the method of stack increases performance by increasing the number of queries. For a fair comparison, our SDQ also collects selective queries from $1^{st}$ and $2^{nd}$ decoder layers. Compared with the other two different methods, our SDQ mechanism is the most effective and is the only one that can reduce the number of queries.

\TableDOTAVonefiveResult
\TableDOTAVtwoResult
\TableResultICDAR

\textbf{Number of decoder layers.}
Multiple decoder layers are a widely used technique to improve detection performance. We performed ablation experiments on different decoder layers on DIOR-R, to further verify the effectiveness of SDQ, shown in Table \ref{ablations} (f). When only one decoder layer, we do SDQ on initial queries. In this condition, our method only achieves 44.02\% mAP. When we increase the number of decoder layers from 2 to 3, the mAP drops from 62.63\% to 62.08\%. The decoder and attention bring too many parameters which may be redundant.

\textbf{Different combinations of selective queries and distinct queries in SDQ.}
To further verify the effectiveness of SDQ, we performed ablation experiments with different combinations of selective queries and distinct queries in SDQ on DIOR-R, shown in Table \ref{ablations} (g). The total number of decoder layers is 3 here. When we collect queries from the $2^{nd}$ decoder layer as selective queries and filter similar queries to obtain distinct queries also at the $2^{nd}$ layer, the model achieves the best 62.08\% mAP on DIOR-R. When selective queries contain more queries from front layers, e.g., initial queries, there is a significant decline in performance. Queries at the front layers can not adequately represent objects.

\textbf{Effects of Individual Strategy.}
We check the effectiveness of each proposed module in the proposed method. In Table \ref{ablations} (h), the baseline detector DDQ-O~\citep{ddq} produces an mAP of 61.66\% on DIOR-R with 300 queries. Sequentially incorporating RRoI attention and SDQ into the baseline detector results in mAP of 62.31\% and 62.41\%, respectively, affirming the effectiveness of each design. When both modules are applied, the model achieves an mAP of 62.76\%. Notably, RRoI attention significantly reduces parameters by 23 million compared to instance interaction in the baseline. Introducing RRoI attention to the baseline model, due to the parameter reduction, increases the FPS by 0.36. Subsequent application of SDQ, requiring the filtering of similar queries, leads to a slight decrease in FPS. Importantly, SDQ brings performance improvement without introducing additional parameters.

\subsection{Main Results}

\textbf{Results on DIOR-R.} As shown in Table \ref{DIOR-R-result}, our proposed RQFormer is compared with CNN-based one-stage and two-stage detectors and transformer-based detectors, including RetinaNet-O~\citep{focal_loss}, DFDet~\citep{DFDet}, SOOD~\citep{SOOD}, Oriented Reppoints~\citep{oriented_reppoints}, DCFL~\citep{dcfl}, Gliding Vertex~\citep{gliding_vertex}, RoI Transformer~\citep{roi_transformer}, QPDet~\citep{QPDet}, AOPG~\citep{dior}, ARS-DETR~\citep{arsdetr}, OrientedFormer~\citep{orientedformer}, and DDQ-O~\citep{ddq}. For fairness comparison, the backbones of all models are ResNet50 in Table \ref{DIOR-R-result}. It achieves the promising performance of 67.31\% mAP with ResNet50 backbone on the DIOR-R benchmark. Our method surpasses all comparison CNN-based one-stage and two-stage detectors and Transformer-based detectors. Specifically, our method outperforms DCFL~\citep{dcfl} by 0.51\% (67.31\% vs. 66.80\%), AOPG~\citep{dior} by 2.90\% (67.31\% vs. 64.41\%), and ARS-DETR~\citep{arsdetr} by 1.41\% (67.31\% vs. 65.90). We reconstruct DDQ-O~\citep{ddq} as our baseline and it achieves an mAP of 66.51\%. Our RQFormer outperforms baseline by 0.80\% (67.31\% vs. 66.51\%). Furthermore, RQFormer performs better in some categories with large aspect ratios, e.g., airport (APO) and dam. Take the category of dam for an example, RQFormer improves Oriented RepPoints~\citep{oriented_reppoints}, AOPG~\citep{dior}, ARS-DETR~\citep{arsdetr} and DDQ-O~\citep{ddq} with the gains of 5.28\%, 8.77\%, 0.95\%, and 6.48\%. It proves that RQFormer can detect accurately objects with extreme aspect ratios.

\TableSDQSparsercnn
\TableIndividualDIORandDOTA
\TableFPS

\textbf{Results on DOTA-v1.0.} We compare our RQFormer method with modern CNN-based one-stage and two-stage detectors and transformer-based detectors, including TIR-Net~\citep{TIR-Net}, R3Det~\citep{r3det}, CFA~\citep{cfa}, SASM~\citep{sasm}, PSC~\citep{psc}, H2RBox~\citep{h2rbox}, SCRDet~\citep{scrdet}, RoI Transformer~\citep{roi_transformer}, Gliding Vertex~\citep{gliding_vertex}, Oriented R-CNN~\citep{orientedrcnn}, Deformable DETR-O~\citep{deformabledetr}, EMO2-DETR~\citep{emo2detr} and ARS-DETR~\citep{arsdetr}. Table \ref{DOTA-1.0-result} reports the detailed results of every category on the DOTA-v1.0 dataset. With ResNet50 as the backbone, RQFormer obtains 75.04\% mAP under single-scale training and testing. In addition, with the multi-scale training augmentation strategy, our method achieves 80.92\% mAP using the ResNet50 backbone. RQFormer surpasses all CNN-based one-stage and transformer-based methods. Specifically, with single-scale data, RQFormer outperforms SASM~\citep{sasm} by 0.12\% (75.04\% vs. 74.92\%) and ARS-DETR~\citep{arsdetr} by 1.25\% (75.04\% vs. 73.79\%). In addition, RQFormer outperforms most CNN-based two-stage methods, except for the Oriented R-CNN~\citep{orientedrcnn}. We argue that it benefits from the complex architecture. Different from these two-stage models, our method does not require hand-craft postprocessing.

\textbf{Results on DOTA-v1.5.}
We report results of small-size objects on DOTA-v1.5 in Table \ref{DOTA-1.5-result}. We compare our method with RetinaNet-O~\citep{focal_loss}, DCFL~\citep{dcfl}, Faster RCNN-O~\citep{faster_rcnn}, Mask R-CNN~\citep{mask_rcnn}, HTC~\citep{htc}, and ReDet~\citep{redet}. The backbones of all models are ResNet50. Our RQFormer performs better for the categories with small instances. Taking small vehicles as an example, our method improves DCFL~\citep{dcfl} and ReDet~\citep{redet} with the mAP gains of 5.44\% and 9.78\%, respectively. It proves the effectiveness of our method on small objects.

\textbf{Results on DOTA-v2.0.}
Table \ref{DOTA-2.0-result} shows a comparison of our method with the modern detectors on the DOTA-v2.0 dataset. We compare our method with SASM~\citep{sasm}, RetinaNet-O~\citep{focal_loss}, Oriented Reppoints~\citep{oriented_reppoints}, Mask R-CNN~\citep{mask_rcnn}, ATSS-O~\citep{atss}, S$^2$A-Net~\citep{s2anet}, HTC~\citep{htc}, DCFL~\citep{dcfl}, RoI Transformer~\citep{roi_transformer}, and Oriented R-CNN~\citep{orientedrcnn}. The backbones of all models are ResNet50 in Table \ref{DOTA-2.0-result}. Our proposed method achieves the competitive performance of 53.28\% mAP under single-scale training and testing. It is comparable to the mAP of the previous best models Oriented R-CNN~\citep{orientedrcnn}.

\textbf{Results on ICDAR2015.} Text detection involves locating text within images using boxes. It is worth noting that, as an extension of RQFormer, we only utilize rotated rectangular boxes to locate texts. We conduct experiments using the ICDAR2015~\citep{icdar2015} dataset, shown in Table~\ref{ResultICDAR2015}. We compare our method with R3DET~\citep{r3det}, GWD~\citep{gwd}, PSC~\citep{psc}, Oriented Reppoints~\citep{oriented_reppoints}, CFA~\citep{cfa}, S$^2$A-Net~\citep{s2anet}, Oriented RCNN~\citep{orientedrcnn}, and DDQ-O~\citep{ddq}. The precision, recall, and F-measure are the evaluation metrics following official criteria~\citep{icdar2015}. We reimplement compared methods using MMRotate~\citep{mmrotate}, with the same settings as the RQFormer. The backbone used by all methods is ResNet50~\citep{resnet}. The proposed RQFormer achieves a precision of 85.0\%, a recall of 75.5\%, and an F-measure of 80.0\%. It enhances the detection performance by 1.0\% compared with the baseline and outperforms all the compared CNN-based models. The inference speed of RQFormer is 14.1 FPS, surpassing the two-stage method Oriented RCNN (10.0 FPS), but lagging behind the one-stage method CFA (21.4 FPS). Due to the limitation of rotating rectangular boxes, oriented object detection methods are not as flexible and accurate as dedicated text detection methods.

\TableQueryNumber
\comparisonsimilarquery
\convergence

\subsection{Generalization to More Tasks}
\textbf{Results on COCO 2017.} To verify the generalization of our method in horizontal object detection, we also conduct experiments on the MS COCO 2017~\citep{coco} dataset. Due to horizontal boxes, RRoI attention is not applicable in this task. Thus, we only apply the Selective Distinct Queries (SDQ) module to the horizontal object detectors Sparse R-CNN~\citep{sparsercnn} and DDQ~\citep{ddq} with the ResNet50 backbone to verify the effectiveness of SDQ. The model is trained and tested on COCO 2017 and detailed results are reported in table \ref{SDQ_result}. In Sparse R-CNN, we collect selective queries only from the $5^{th}$ decoder layer out of a total of 6 decoder layers, while in DDQ, we collect them from the $1^{st}$ decoder layer out of a total of 2 decoder layers. We then obtain distinct queries before the final decoder layers, with the IoU threshold set to 0.7. With 100 queries and 12 training epochs, SDQ significantly improves mAP by +3.1\% compared to Sparse R-CNN and by +0.3\% compared to DDQ. Increasing the number of queries to 300 and training epoch to 36, SDQ achieves 46.8\% (+1.8\%) mAP on the Sparse R-CNN~\citep{sparsercnn} with the help of a random resize augmentation strategy~\citep{detr}. These experiments can fully prove the effectiveness of the SDQ method. The SDQ would not bring extra parameters of the model and FPS is only slightly reduced (20.0 vs. 20.6).

\subsection{Effectiveness of individual modules}
To further investigate the effectiveness of the individual proposed module, we gradually integrate RRoI attention and SDQ into the baseline DDQ-O~\citep{ddq}, training them on both DIOR-R and DOTA-v1.0. Model evaluation involves mAP, parameters, and FPS, shown in Table~\ref{idividual_on_dior_dota1.0}. In this experiment, the number of queries is set to 500, and models are trained for 36 epochs on DIOR-R and 24 epochs on DOTA-v1.0. The FPS is tested on an RTX 2080ti. The baseline DDQ-O achieves a mAP of 66.51\% and 73.57\% on DIOR-R and DOTA-v1.0, respectively. When only incorporating RRoI attention, models exhibit an mAP improvement of 0.5\% on DIOR-R and 0.48\% on DOTA-v1.0. RRoI attention significantly reduces model parameters (40.93M vs. 63.73M on DIOR-R), as it focuses only on local areas within boxes. Fewer parameters result in a slight increase in FPS. With the integration of only SDQ, models demonstrate an mAP improvement of 0.76\% on DIOR-R and 1.32\% on DOTA-v1.0. SDQ does not introduce extra parameters to models, benefiting from its capability to filter similar queries. In this condition, the FPS experiences a slight decrease since SDQ first collects queries from intermediate layers and then filters similar queries. When both RRoI attention and SDQ are applied on the baseline, RQFormer achieves a mAP of 67.31\% and 75.04\% on DIOR-R and DOTA-v1.0, respectively.

\subsection{Comparison of speed, parameters and accuracy}
To further assess the performance of RQFormer, we conducted a comparison with other methods in terms of FPS, parameters, and accuracy on the DOTA-v1.0 dataset, shown in Table~\ref{FPS_Params}. RQFormer achieved a detection speed of 11.20 FPS, surpassing all compared CNN-based two-stage methods, but trailing behind most one-stage methods. We attribute the slower speed of CNN-based two-stage models to their complex structure and the influence of handcrafted post-processing modules. On the other hand, CNN-based one-stage models, characterized by a simpler framework and strategy than end-to-end models, tend to be faster. Notably, the speed of RQFormer is faster than all compared end-to-end methods, which we attribute to the elimination of an encoder and fewer decoder layers in our approach. The parameters of RQFormer amount to 40.90M, significantly less than baseline DDQ-O~\citep{ddq} (63.71M), while achieving a higher mAP. The parameters of RQFormer are also fewer than most compared CNN-based two-stage and end-to-end methods, though slightly larger than most one-stage methods. It proves the efficiency of our method in terms of parameters.

\subsection{Effects of Queries}
We performed ablation experiments with different number of queries, which including positional and content queries, as shown in Table~\ref{query_number}. The number of positional queries and content queries is always equal. As the number of queries increases from 100 to 500, the mAP improves from 60.23\% to 67.31\%. Since remote sensing images contain a large number of densely packed objects, having a sufficient number of queries allows for better coverage of these objects. The number of parameters in the model only increases slightly, by 0.1M.

\queries
\DOTAvisual

\subsection{Comparison of similar queries}
We calculate the ratios of similar queries for our method and other end-to-end methods on DOTA-v1.0 to further validate the effectiveness of SDQ. The formula is shown as Eq.~\ref{eq_sdq} and results are illustrated in Fig.~\ref{comparison_of_similar_query}. Our approach effectively filters out similar queries, resulting in a notable reduction in their ratio across four different IoU threshold settings: 0.8, 0.85, 0.9, and 0.95. Taking a threshold of 0.95 as an example, the ratios of similar queries are 15.20\%, 9.63\%, 10.16\%, and 3.86\% in Deformable DETR-O, Deformable DETR-O with CSL, ARS-DETR, and DDQ-O, respectively. Our method can significantly reduce it to 0.79\%, which serves as evidence of its effectiveness.

\subsection{Convergence}
DETR~\citep{detr} suffers from long training time and slow convergence. The Deformable DETR series~\citep{deformabledetr, arsdetr} methods alleviate this issue. To further investigate training efficiency, we compare RQFormer with other end-to-end models. Convergence curves for these models are depicted in Fig.~\ref{convergence_curve}. For a fair comparison, all methods are trained on 12 epochs on DIOR-R using ResNet50 with 300 queries. The learning rate decays by 10$\times$ on the $8^{th}$ and $11^{th}$ epochs respectively. RQFormer achieves a mAP of 62.8\% in just 12 epochs, surpassing Deformable DETR-O with CSL~\citep{csl} (31.23\%) and ARS-DETR~\citep{arsdetr} (38.9\%). Additionally, RQFormer outperforms baseline DDQ-O by 1.1\%. These results underscore the effectiveness of our approach in mitigating the challenges associated with slow convergence.

\subsection{Visualization}
\textbf{Visualization of queries through Selective Distinct Queries.}
Shown in Fig. \ref{queries}, we visualize initial queries, selective queries, distinct queries, and final queries. Initial queries are more random. There are many highly overlapping boxes. Selective queries are more accurate than initial queries. They effectively cover objects. We filter similar queries to obtain distinct queries. Boxes with high overlapping are filtered out. Final queries can detect objects accurately.

\comparewithothermethods
\DIORvisual
\ICDARvisual

\textbf{Visualization of RRoI attention.}
We visualize RRoI attention in Fig. \ref{dota_visual} without using CAM. We observe that it mostly focuses on object extremities such as the front and back of planes and ships. These highlighted areas are very similar to DETR~\citep{detr} and Deformable DETR~\citep{deformabledetr}.

\textbf{Comparison with other methods.} We compare our method with others on large aspect ratio objects, densely packed objects, and square-like objects, as shown in Fig.~\ref{compare_with_other_method}. Other methods failed to detect these challenging objects.

\textbf{Visualization of results.}
We visualize some predicted results on DOTA-v1.0 in Fig. \ref{dota_visual}, DIOR-R in Fig. \ref{dior_visual}, and ICDAR2015 in Fig.~\ref{icdar2015_visual}. We can see that our model detects objects accurately both in positions and categories.

\section{Conclusions}
In this paper, we propose an end-to-end detector RQFormer for oriented object detection that incorporates two key technologies: Rotated RoI Attention (RRoI Attention) and Selective Distinct Queries (SDQ). The proposed RRoI Attention mechanism aligns keys and positional queries through cross-attention, effectively handling objects with arbitrary orientations and varying scales. To prevent similar queries from hindering optimization, SDQ first collects high-quality selective queries from intermediate decoder layers and then filters out similar ones, ensuring that the remaining queries are distinct. This process facilitates one-to-one label assignment. Extensive experiments demonstrate that our method achieves competitive accuracy compared to modern CNN-based one-stage and two-stage detectors, as well as transformer-based detectors. We hope our approach will inspire researchers to further explore the application of end-to-end models in oriented object detection.

\textbf{Limitations and feature works}. The Rotated RoI attention primarily focuses on local areas strictly within positional queries. Contextual information around these boxes may enhance attention. Further exploration into integrating contextual information into attention is necessary.

\section*{CRediT authorship contribution statement}
Jiaqi Zhao: Conceptualization, Methodology, Software, Writing – original draft, Writing – review \& editing. Zeyu Ding: Conceptualization, Methodology, Software, Writing – original draft, Writing – review \& editing. Yong Zhou: Visualization, Formal analysis, Funding acquisition, Supervision. Hancheng Zhu: Data curation, Investigation, Writing – review \& editing. Wen-Liang Du: Data curation, Investigation, Writing – review \& editing. Rui Yao: Software, Validation. Abdulmotaleb El Saddik: Validation, Writing – review \& editing.
\section*{Declaration of competing interest}
The authors declare that they have no known competing financial
interests or personal relationships that could have appeared to influence
the work reported in this paper. 
\section*{Data availability}
Data will be made available on request. 
\section*{Acknowledgments}
This work was supported by the National Natural Science Foundation of China (No.62272461,62002360, 62172417, 62101555), the Double FirstClass Project of China University of Mining and Technology for Independent Innovation and Social Service under Grant 2022ZZCX06, the Six Talent Peaks Project in Jiangsu Province (No. 2015-DZXX-010, 2018-XYDXX-044).

\bibliographystyle{cas-model2-names}

\bibliography{reference}

\end{sloppypar}
\end{document}